\documentclass{IEEEtran}

\usepackage[numbers]{natbib}
\usepackage[utf8]{inputenc}
\usepackage{amsmath,amssymb,amsfonts,mathtools,bm}
\usepackage[scosf]{newtx} 
\usepackage{graphicx,xcolor,blox}
\definecolor{MatBlue}{rgb}{0, 0.4470, 0.7410}
\definecolor{MatOrange}{rgb}{0.8500, 0.3250, 0.0980}
\definecolor{MatYellow}{rgb}{0.9290, 0.6940, 0.1250}
\usepackage{tikz}
\usetikzlibrary{calc,fit,arrows,arrows.meta,positioning,backgrounds}
\usepackage{dblfloatfix}

\usepackage{textcomp}
\usepackage{siunitx}
\usepackage{hyperref}
\usepackage{tabularx}
\usepackage{float}

\usepackage{algorithm}
\usepackage[noend]{algpseudocode}
\usepackage{booktabs,paralist}

\usepackage[
placement=top,
angle=0,
color=red,
scale=1.1,
hshift=0,
vshift=-15
]{background}
\backgroundsetup{contents=\bf{----------------------------------------------------------------- PREPRINT -----------------------------------------------------------------}}

\def\BibTeX{{\rm B\kern-.05em{\sc i\kern-.025em b}\kern-.08em
    T\kern-.1667em\lower.7ex\hbox{E}\kern-.125emX}}
\ifCLASSOPTIONcompsoc
\usepackage[caption=false,font=normalsize,labelfont=sf,textfont=sf]{subfig}
\else
\usepackage[caption=true,font=footnotesize]{subfig}
\fi

\newcommand{\diff}[1]{\ensuremath{\operatorname{d}\!{#1}}}
\DeclareMathOperator{\sign}{sign}
\newcommand{\eqdef}{=\vcentcolon}
\newcommand{\defeq}{\vcentcolon=}
\newcommand{\vect}[1]{\ensuremath{\textbf{#1}}}
\DeclareMathOperator*{\argmax}{arg\,max}   
\DeclareMathOperator*{\atantwo}{atan2}   

\newcommand{\steerangle}{\ensuremath{\delta}}
\newcommand{\steeranglelin}{\ensuremath{\tilde{\delta}}}
\newcommand{\wheelbase}{\ensuremath{\ell}}
\newcommand{\track}{\ensuremath{w}}
\newcommand{\curvature}{\ensuremath{\kappa}}
\newcommand{\head}{\ensuremath{\psi}}
\newcommand{\vel}{\ensuremath{v}}
\newcommand{\velmax}{\ensuremath{{v^{\text{\scshape max}}}}}
\newcommand{\latacc}{\ensuremath{a_c}}
\newcommand{\lataccmax}{\ensuremath{{\latacc^{\text{\scshape max}}}}}
\newcommand{\dt}{\ensuremath{\diff{t}}}
\newcommand{\pathradius}{\ensuremath{{\rho_s}}}
\newcommand{\headerr}{\ensuremath{e_{\head}}}
\newcommand{\laterr}{\ensuremath{e_{y}}}
\newcommand{\dotheaderr}{\ensuremath{\dot{e}_{\head}}}
\newcommand{\dotlaterr}{\ensuremath{\dot{e}_{y}}}
\newcommand{\state}{\vect{x}}
\newcommand{\outtput}{\vect{y}}
\newcommand{\innput}{\vect{u}}
\newcommand{\dotx}{\ensuremath{\dot{\state}}}
\newcommand{\lookahead}{\ensuremath{{L_d}}}
\newcommand{\lle}{\ensuremath{e_d}}
\newcommand{\lhe}{\ensuremath{\alpha}}
\newcommand{\steerdelay}{\ensuremath{\tau_{d}}}
\newcommand{\steerlag}{\ensuremath{\tau}}
\newcommand{\ppgainprop}{\ensuremath{{K^P_{pp}}}}
\newcommand{\ppgainder}{\ensuremath{{K^D_{pp}}}}
\newcommand{\ppgainderPD}{\ensuremath{{K^{D^\star}_{pp}}}}
\newcommand{\jw}{\ensuremath{j\omega}}
\newcommand{\w}{\ensuremath{\omega}}
\newcommand{\wsq}{\ensuremath{\eta}}
\newcommand{\laterrmax}{ \ensuremath{ \laterr^{\text{\scshape max}} } }
\newcommand{\headerrmax}{ \ensuremath{ \headerr^{\text{\scshape max}} } }

\begin{document}
\setlength{\parindent}{0pt}

\title{A Sim-to-Real Vision-based Lane Keeping System for a 1:10-scale Autonomous Vehicle}

\author{
Antonio Gallina, Matteo Grandin, Angelo Cenedese, \IEEEmembership{Member, IEEE}, Mattia Bruschetta, \IEEEmembership{Member, IEEE}
\thanks{Antonio Gallina, Angelo Cenedese and Mattia Bruschetta are with the Department of Information Engineering, University of Padua, 35131 Padua, Italy (e-mail: \texttt{\{antonio.gallina, angelo.cenedese, mattia.bruschetta\}@dei.unipd.it}).}
\thanks{Matteo Grandin is with the Centro Ricerche Fusione, University of Padua, 35127 Padua, and the Department of Information Engineering, University of Padua, 35131 Padua, Italy (e-mail: \texttt{matteo.grandin@phd.unipd.it})}} 


\maketitle

\begin{abstract}

In recent years, several competitions have highlighted the need to investigate vision-based solutions to address scenarios with functional insufficiencies in perception, world modeling and localization.
This article presents the Vision-based Lane Keeping System (VbLKS) developed by the DEI-Unipd Team within the context of the Bosch Future Mobility Challenge 2022. The main contribution lies in a Simulation-to-Reality (Sim2Real) GPS-denied VbLKS  for a 1:10-scale autonomous vehicle. In this VbLKS, the input to a tailored Pure Pursuit (PP) based control strategy, namely the Lookahead Heading Error (LHE), is estimated at a constant lookahead distance employing a Convolutional Neural Network (CNN). 
A training strategy for a compact CNN is proposed, emphasizing data generation and augmentation on simulated camera images from a 3D Gazebo simulator, and enabling real-time operation on low-level hardware.
A tailored  PP-based lateral controller equipped with a derivative action and a PP-based velocity reference generation are implemented. Tuning ranges are established through a systematic time-delay stability analysis.
Validation in a representative controlled laboratory setting is provided.
\end{abstract}

\begin{IEEEkeywords}
autonomous vehicle, lane keeping, sim-to-real, vision-based, gps-denied, pure pursuit, competition
\end{IEEEkeywords}

\section{Introduction}\label{sec:introduction}
\begin{figure}[!t]
  \centering
  \includegraphics[width=.99\columnwidth]{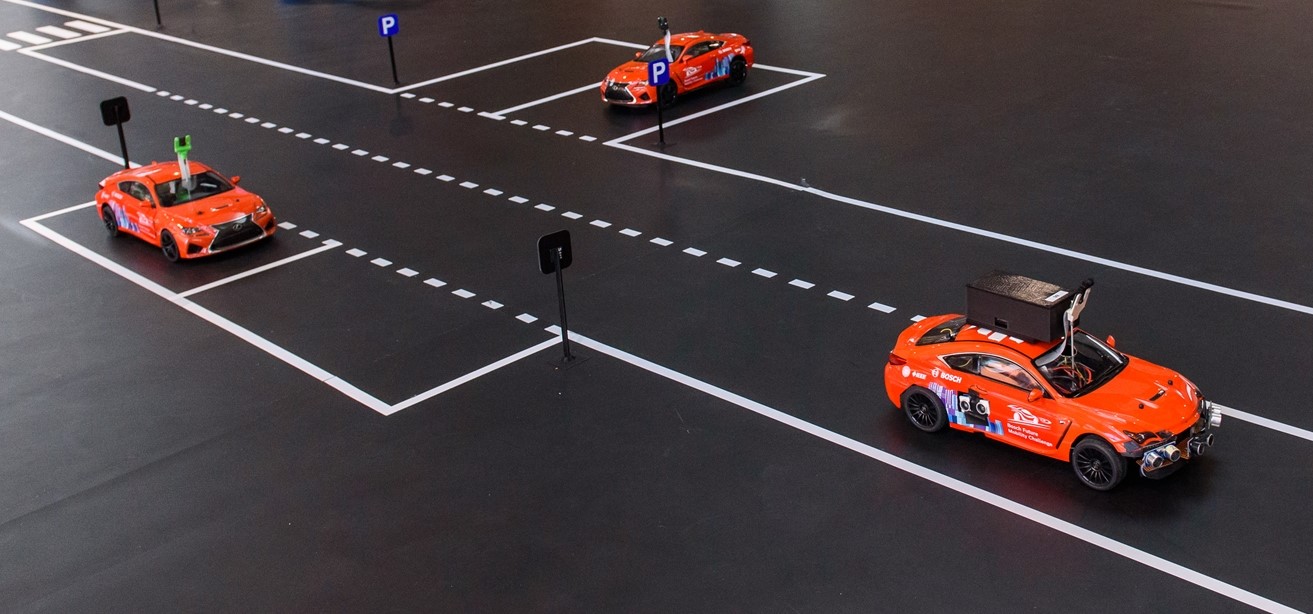}
	\caption{DEI-Unipd Team's vehicle during BFMC 2022}
	\label{fig:bfmc-dei-car}
\end{figure}

Research on Autonomous Vehicles (AVs) has experienced an increasingly significant growth of interest in the last few years because of its potential to enhance safety, efficiency, and convenience of automotive transportation. Anyway, due to its complexity, there are still many technical and social challenges to be tackled in this ﬁeld~\cite{Chen2022}.
Driving tasks have become increasingly complex in recent years, with the advent of new technologies and the need for higher levels of automation.
Among the driving tasks a full self-driving vehicle~\cite{SAEJ3016} should be capable of performing, a key role is played by the Lane Keeping System (LKS), which is responsible for driving the vehicle within its lane on the road.
Over the past few years, several competitions have played a crucial role in advancing the development of Autonomous Driving (AD) algorithms, serving as a test bed for various LKS implementations tailored specifically for embedded hardware. Notable competitions in this domain include the DARPA Urban Challenge~\cite{DARPA2009,Campbell2007}, the Duckietown AI Driving Olympics~\cite{Kalapos2020,Almasi2020}, the Carolo Cup~\cite{Cars1,Cars2}, and the Mexican Robotics Tournament~\cite{Cars3}. 
Remarkably, the importance of using the Simulation-to-Reality (Sim2Real) framework is recently emerging intending to enable the development of complex AD algorithms while reducing implementation costs and time to access experimental scenarios. Nonetheless, challenges persist, notably in the form of disparities between simulated environments and the real world~\cite{Hu2023}.
The Bosch Future Mobility Challenge (BFMC), hosted by the Bosch Engineering Center in Cluj (RO), represents a recent addition to this landscape, further strengthened by its collaboration with IEEE ITSS since 2021~\cite{BFMC-ITSS}. This international technical competition invites teams of students to develop autonomous driving algorithms on 1:10 scale vehicles, in an environment that mimics a miniature smart city (see Fig.~\ref{fig:bfmc-dei-car}).
This article presents the LKS developed by the DEI-Unipd Team within the context of the BFMC 2022, showcasing its pivotal role in the team's victory.

\subsection{Context and state of the art} 
\begin{figure}[!ht]
    \centering
    \begin{tikzpicture}[>=latex,node distance=.5cm]
        \footnotesize
        \tikzstyle{lksnode} =[draw, fill=MatOrange!15, inner sep=5pt,outer sep=0pt]
        \tikzstyle{vblksnode} =[draw, fill=MatBlue!15, inner sep=5pt,outer sep=0pt]
        \tikzstyle{e2elksnode} =[draw, fill=MatYellow!15, inner sep=5pt,outer sep=0pt]
        
        \node[lksnode] (sensing) {\begin{minipage}{\widthof{\& Perception}}
            \centering
            Sensing\\\& Perception
        \end{minipage}};
        \node[lksnode,right= .35cm of sensing] (planning) {\begin{minipage}{\widthof{Planning}}
            \centering
            Path\\Planning
        \end{minipage}};

        \coordinate [below= .8cm of sensing.south west] (a);
        \coordinate [below= .8cm of planning.south east] (b);
        \node [vblksnode, fit={(a) (b)}, minimum height = 1cm, inner sep=0pt, label=center:\begin{minipage}{\widthof{Integrated Perception}}
            \centering
            Integrated Perception\\\& Planning
        \end{minipage}] (integrated) {};

        \newlength{\tmplgth}
        \setlength{\tmplgth}{\widthof{Tracking}}
        \coordinate [right= 1.1cm of planning.north east] (c);
        \coordinate [right= 1.1cm of integrated.south east] (d);
        \node[lksnode, fit={(c) (d)}, minimum width = 1.5\tmplgth, inner sep=0pt, label=center:
        \begin{minipage}{\tmplgth}
            \centering
            Path\\Tracking
        \end{minipage}] (tracking) {};

        \coordinate [below= of integrated.south west] (e);
        \coordinate [below= of tracking.south east] (f);
        \node [e2elksnode, fit={(e) (f)}, minimum height = .5cm, inner sep=0pt, label=center:End-to-End Learning] (endtoend) {};

        \draw[->] (sensing) -- (planning);
        \draw[->] (planning) -- (planning-|tracking.west);
        \draw[->] (integrated) -- (integrated-|tracking.west);
        
        \begin{pgfonlayer}{background}
            \node [fill=black!10,fit= {(sensing) (endtoend)}, inner sep=.75em] (lks) {};
        \end{pgfonlayer}

        \node[left = .5cm of sensing] (input) {\begin{minipage}{\widthof{sensor}}
            \centering\itshape\scriptsize
            sensor\\output
        \end{minipage}};
        \node[right = .6cm of tracking] (output) {\begin{minipage}{\widthof{control}}
            \centering\itshape\scriptsize
            control\\input
        \end{minipage}};

        \draw[->] (input) -- (sensing);
        \draw[->] (input.east) -- ($(input.east)!0.4!(sensing.west)$) |- (integrated.west);
        \draw[->] (input.east) -- ($(input.east)!0.4!(sensing.west)$) |- (endtoend.west);;

        \draw[->] (tracking.east) |- (output);
        \draw[->] (endtoend.east) -| ($(tracking.west)!0.87!(output.west)$) -- (output.west);;

    \end{tikzpicture}
    \caption{Block scheme of the Lane Keeping System architecture}
    \label{fig:lks-block}
\end{figure}
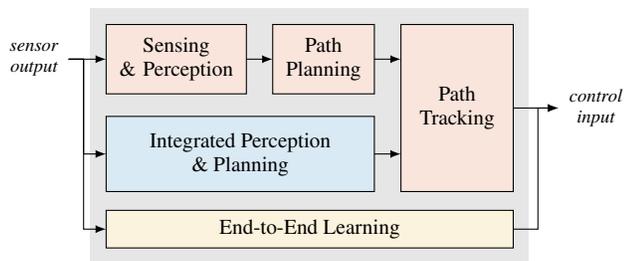

The architecture of a LKS (Fig.~\ref{fig:lks-block}) can be mainly divided into three sequential layers, namely \emph{Sensing \& Perception}, \emph{Path Planning}, and \emph{Path Tracking}.

The \emph{Sensing \& Perception} layer is responsible for sensing the environment around the vehicle and detecting the relative positioning of the vehicle concerning lane markings on the road. It includes various sensors such as cameras, Global Positioning Systems (GPSs), lidars, and radars that capture accurate and reliable data and convert it into a digital format that can be processed by the next level.

Once the Perception component has gathered the necessary data, the \emph{Path Planning} layer uses this information to determine the appropriate path the vehicle should follow to stay within its lane based on the current driving situation.

Finally, the \emph{Path Tracking} layer translates the output from the Path Planning component into suitable signals to drive steering and speed actuators for controlling vehicle's lateral and longitudinal dynamics, ensuring smooth and safe driving.
Specifically focusing on this latter layer, the ﬁeld of automatic steering control is fairly well established~\cite{Paden2016,Rokonuzzaman2021} and several Path Tracking Controllers (PTCs) have been employed in low-level embedded solutions, ranging from simple LUT~\cite{Kumar2022}, PID~\cite{Diab2020a,Cantas2018}, and Geometric Control~\cite{Kuo2019,Htet2015}, up to more complex optimization-based techniques such as NMPC~\cite{Satria2022}.

An alternative to the previously described sequential framework (see Fig.~\ref{fig:lks-block}) involves the adoption of various motion planning techniques within an integrated approach, allowing the retrieval of a feasible path directly from sensor output data (\emph{Integrated Perception and Planning})~\cite{Schwarting2018}. Additionally, more recently, different end-to-end solutions employ complex Deep Learning (DL) architectures to estimate the control input directly from sensor output data (\emph{End-to-End Learning})~\cite{Bojarski2017,Gidado2020,Kocic2021}.

Within the LKS architecture summarized in Fig.~\ref{fig:lks-block}, Vision-based LKSs (VbLKSs) have gained prominence in recent years owing to the effectiveness of cameras in providing comprehensive information content without disrupting other sensors, while simultaneously paving the way for GPS-denied solutions to the lane keeping task.
In the VbLKS scenario, Perception is usually attained through Lane Detection methods, which process camera frames in order to identify lane markings on the road ahead~\cite{Yenikaya2013}.
According to the literature, Lane Detection techniques can be classified into two major categories: classical Computer Vision (CV) techniques and Machine Learning (ML) approaches.
Classical CV techniques employ image processing methods such as Canny edge detection and Hough transform, to extract road information from the output camera frame~\cite{Xing2018, Huang2021}. Despite being computationally efficient, these methods require complex post-processing techniques, namely Lane Tracking, to address their inherent lack of robustness: Kalman or Particle filters are usually employed to provide smoother signals~\cite{Zakaria2022, Waykole2021}.
More recently, to improve the overall detection and recognition accuracy and robustness, different ML-based techniques have been proposed: in turn, these methods require a large amount of unbiased training data along with increasing computational burden, making them unsuitable for embedded applications~\cite{Sarmiento2022}.

\subsection{Contribution of the article}
\begin{figure}[H]
    \centering
    \begin{tikzpicture}[>=latex,node distance=.75cm]
        \small
        \tikzstyle{lksnode} =[draw, fill=MatOrange!15, inner sep=5pt,outer sep=0pt]
        \tikzstyle{vblksnode} =[draw, fill=MatBlue!15, inner sep=5pt,outer sep=0pt]
        \tikzstyle{e2elksnode} =[draw, fill=MatYellow!15, inner sep=5pt,outer sep=0pt]

        \node[] (input) {\begin{minipage}{\widthof{camera}}
            \centering\itshape
            camera\\frame
        \end{minipage}};
        
        \node[vblksnode, right = of input] (estimation) {\begin{minipage}{\widthof{estimation}}
            \centering
            CNN\\estimation
        \end{minipage}};

        \node[draw,circle,fill,inner sep=.75pt, right = of estimation.east] (derivation) {};

        \node[lksnode, above right = .1cm and .5cm of derivation] (pp-pd) {\begin{minipage}{\widthof{controller}}
            \centering
            PP-D\\controller
        \end{minipage}};
        \node[right = of pp-pd] (steering) {$\steerangle$};

        \node[lksnode, below right = .1cm and .5cm of derivation] (pp-vr) {\begin{minipage}{\widthof{generator}}
            \centering
            PP-VR\\generator
        \end{minipage}};
        \node[right = of pp-vr] (velocity) {$\vel$};

        \draw[-] (estimation) -- node[above]{\textsc{lhe}} (derivation);
        \draw[->] (derivation) |- (pp-pd);
        \draw[->] (derivation) |- (pp-vr);
        \draw[->] (input) -- (estimation);
        \draw[->] (pp-pd) -- (steering);
        \draw[->] (pp-vr) -- (velocity);
        
        \begin{pgfonlayer}{background}
            \node [fill=black!10,fit= {(estimation) (pp-pd) (pp-vr)}, inner sep=.75em] (lks) {};
        \end{pgfonlayer}

\end{tikzpicture}
    \caption{Block scheme of the proposed VbLKS}
    \label{fig:vblks-block-scheme}
\end{figure}
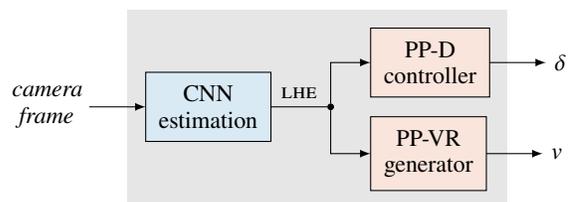

In this article we present a novel Sim2Real GPS-denied VbLKS for a 1:10-scale autonomous vehicle (Fig.~\ref{fig:vblks-block-scheme}).
The VbLKS architecture proposed here hinges on the well-known Pure Pursuit (PP) strategy~\cite{Coulter1992,Rokonuzzaman2021} and is divided into two parts according to the integrated framework depicted in Fig.~\ref{fig:lks-block}.
Within this framework, the input to the Path Tracking block, hereinafter referred to as Lookahead Heading Error (LHE), is computed at a constant lookahead distance by means of a tailored Convolutional Neural Network (CNN).
Following the domain randomization approach in the Sim2Real framework~\cite{Hu2023}, the CNN is trained solely on simulated camera images coming from a 3D Gazebo open-source simulator mimicking the real-world reference scenario.
To this end, we propose also a tailored training strategy that includes generating and augmenting a considerable amount of data, leveraging on the simulation framework, which allows for generalizing the approach to the real-world scenario.
Specifically focusing on performance, the CNN is characterized by a compact architecture making it able to run in real-time on low-level hardware~\cite{Iandola2017}.

The second main contribution is framed into the Path Tracking block and comprises both a tailored PP path tracking control strategy for lateral control (\textit{PP-D controller}) and a PP-based velocity reference generation technique (\textit{PP-VR generator}) for the longitudinal one. Leveraging on the smooth estimates coming from the CNN, a derivative action added to the standard PP PTC is proposed to cope both with the constraint of fixed lookahead distance and with the input delay introduced by the steering actuation mechanism~\cite{Wang2022}. In addition to this, significant tuning ranges are derived, founded on a systematic time-delay stability analysis.
Eventually, still based on the LHE, we introduce a PP-based control strategy for generating velocity references that aims at maximizing speed while constraining lateral acceleration.

The joint combination of the CNN, the PP-D controller, and the PP-VR generator allows for a GPS-denied solution to the lane-keeping task.
The efficacy of the proposed VbLKS has been validated in a controlled laboratory setting and demonstrated particular proficiency in the well-structured BFMC reference scenario.


The remainder of this paper is organized as follows. Sec.~\ref{sec:refscen-geomform} outlines the BFMC reference scenario and the geometric formulation underlying the proposed approach. Sec.~\ref{sec:cnn-estimation} characterizes the proposed CNN estimation part, along with its training strategy in the Sim2Real framework. Sec.~\ref{sec:pp-control} covers the PP-based control strategies and a time-delay stability analysis of the closed-loop system. Eventually in Sec.~\ref{sec:results}, we report significant results in a structured laboratory scenario. 





\begin{figure*}[b]
    \subfloat[Simulated 3D environment]{\includegraphics[width=\columnwidth]{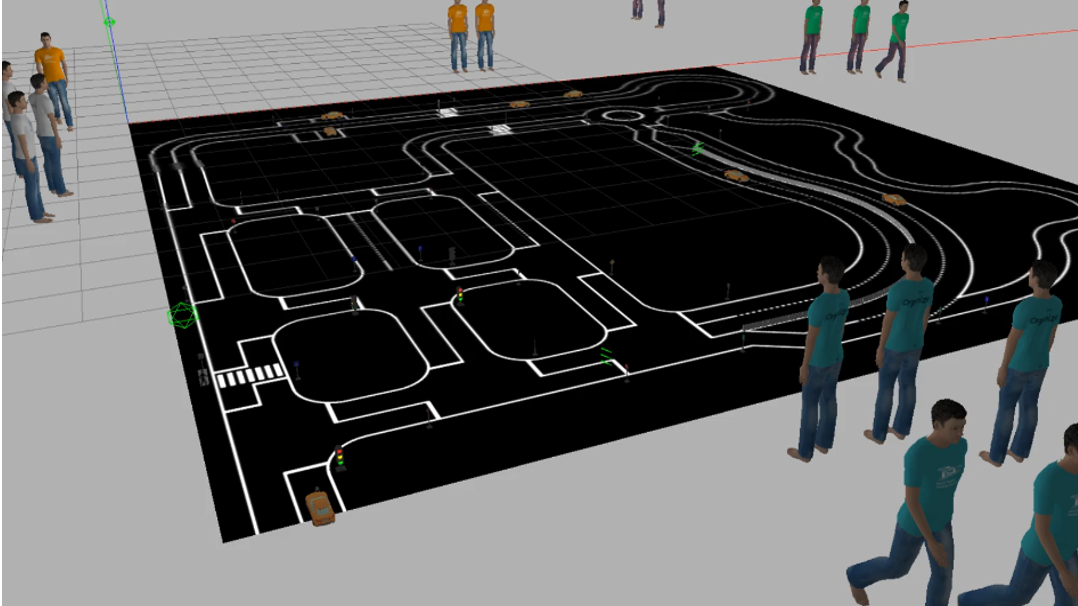}
    		\label{fig:bfmc_sim}}
    \hspace{\columnsep}%
    \subfloat[Real miniature smart city]{\includegraphics[width=\columnwidth]{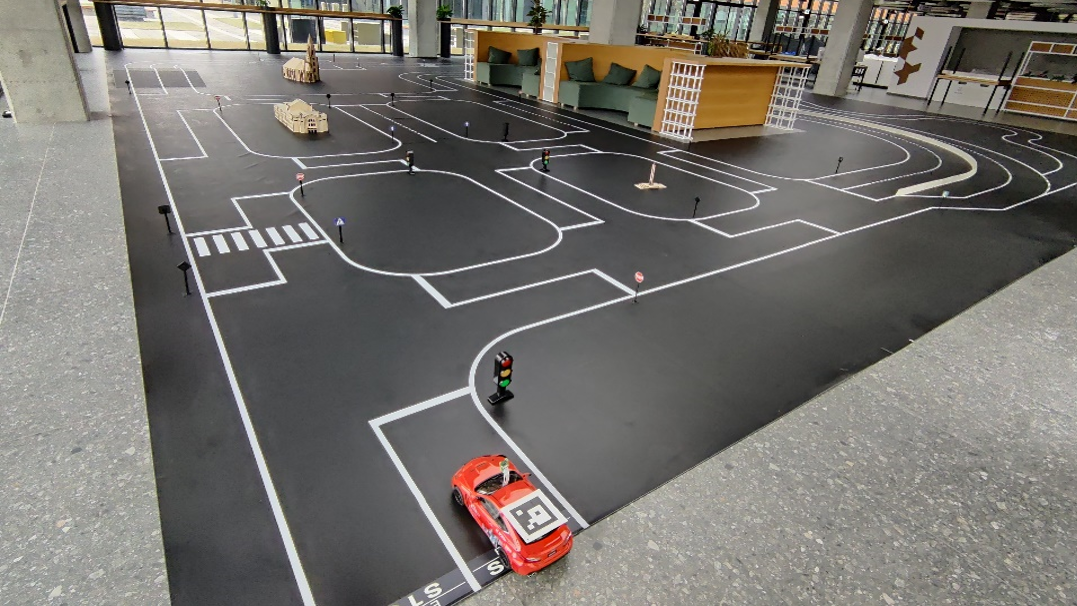}
    		\label{fig:bfmc_real}}
    \caption{Bosch Future Mobility Challenge 2022 scenario}
    \label{fig:bfmc_environment}
\end{figure*}

\section{Reference Scenario and Geometric Formulation} \label{sec:refscen-geomform}
This section provides a concise overview of the BFMC reference scenario, highlighting the key characteristics that significantly influence the design of the control architecture.

The BFMC 2022 comprises two distinct tasks: a \textit{technical run} and a \textit{speed run}. In the technical challenge, AVs are tasked with navigating the map avoiding designated obstacles, performing different maneuvers, and adhering to traffic rules with minimal errors. Conversely, the speed challenge requires AVs to swiftly reach a specified point on the map regardless traffic signs, with no obstacles, and within a limited time frame.
Both tasks necessitate an LKS capable of running on low-level hardware, resilient to real-world non-idealities, and able to retain tracking performance as the vehicle speed increases.

\subsection{Reference scenario}\label{subsec:reference-scenario}
The reference scenario of the BFMC 2022 representing a miniature smart city is illustrated in Fig.~\ref{fig:bfmc_environment}. A significant constraint in this work is represented by the impossibility of accessing the real environment up to the very final stage of the competition. To address this limitation, along with the 1:10 scale vehicle, participants are provided with a 3D Gazebo open-source simulator, replicating the entire BFMC infrastructure (Fig.~\ref{fig:bfmc_sim}).

\subsubsection{Environmental scenario}
Focusing on the environmental scenario, the competition track is composed of straight sections, small curves with a radius equal to $0.65\,\si{\meter}$, and larger curves with a radius equal to $1.04\,\si{\meter}$. All these sections are characterized by a lane width equal to $0.37\,\si{\meter}$.
The inability to access the physical facility hinders the possibility of gathering real-world data. On the other hand, the 3D Gazebo simulated environment allows the generation of a limitless amount of unbiased simulated data. Besides reproducing all the competition obstacles, this simulated environment does not reproduce nonidealities characteristic of a real-world scenario such as surface textures, lighting conditions, and precise vehicle dynamics, suggesting the need for a Sim2Real-based solution able to fill this reality gap. 
Concerning the sensor aspect, the smart city is equipped with a localization system that emulates a real GPS, operating at a frequency of approximately $4\,\si{Hz}$. However, this system introduces a static delay of about $0.4\,\si{\second}$, a standard deviation of about $10\,\si{cm}$, and includes some uncovered areas, making it unreliable for precise motion tracking techniques like optimal model-based control strategies. 



\subsubsection{Hardware setup}
\begin{figure}[!ht]
	\centering
	\includegraphics[width=.8\columnwidth]{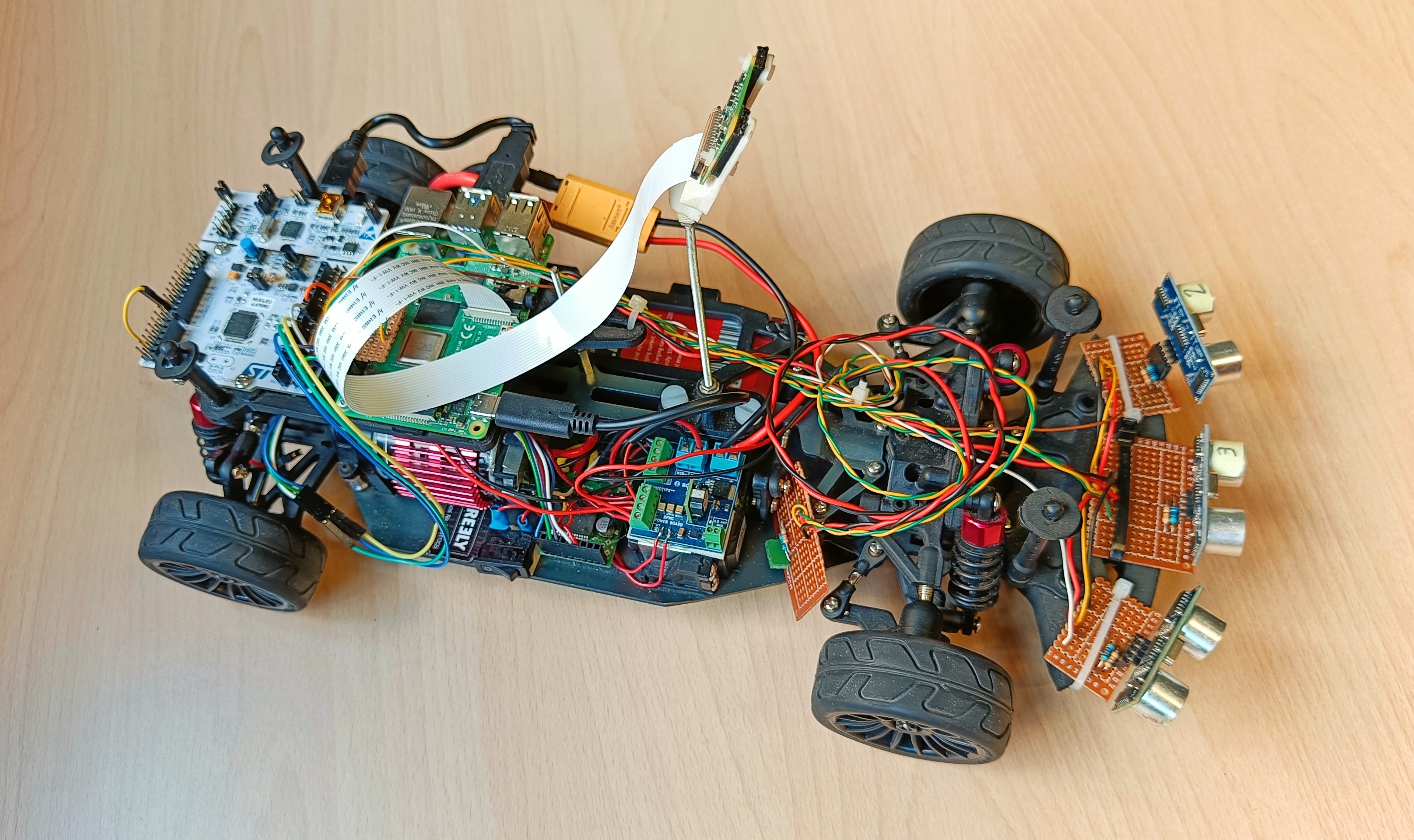}
	\caption{DEI-Unipd Team's 1:10 scale vehicle}
	\label{fig:top-car}
\end{figure}

On the vehicle side (Fig.~\ref{fig:top-car}), the central core is represented by a Raspberry Pi 4 Model B (RPi), having a Broadcom \textsc{bcm2711}, Quad-core Cortex-A72 64-bit SoC running at up to 1.5 GHz. 
The actuation part is handled by a Nucleo \textsc{stm32l476rg} board that forwards speed and steering commands from the RPi to a DC motor through a \textsc{vnh5019} power driver and to a Reely \textsc{rs-610wpmg} analog DC servomotor, respectively. The latter is coupled with a low-level steering transmission that introduces bounds and slack in steering maneuvers. Additionally, steering maneuvers are characterized also by a static delay of about $0.15\,\si{\second}$ and dynamics with a time constant of about $0.17\,\si{\second}$, making high-speed maneuvers challenging.

Above the vehicle, a RasPi Camera V2 module is fixed and connected to the RPi where images are acquired with a resolution of $640\times 480$ px at $30$ FPS, thus enabling a vision-based approach to the lane-keeping task. Nevertheless, the presence of low-level embedded hardware restricts the adoption of state-of-the-art DL techniques.
The orientation of the vehicle is measured by means of a Bosch \textsc{bno055} Smart IMU sensor connected to the RPi via I\textsuperscript{2}C protocol. Speed and relative position feedback is retrieved by an \textsc{amt103} high-accuracy incremental encoder mounted on the shaft of the DC motor.
Everything is powered by a Li-Ion battery pack, interfaced with peripherals through a \textsc{dc-dc} converter.


\subsection{Geometric formulation}\label{subsec:geomform}
\begin{figure}[!ht]
	\centering
	\includegraphics[width=.9\columnwidth]{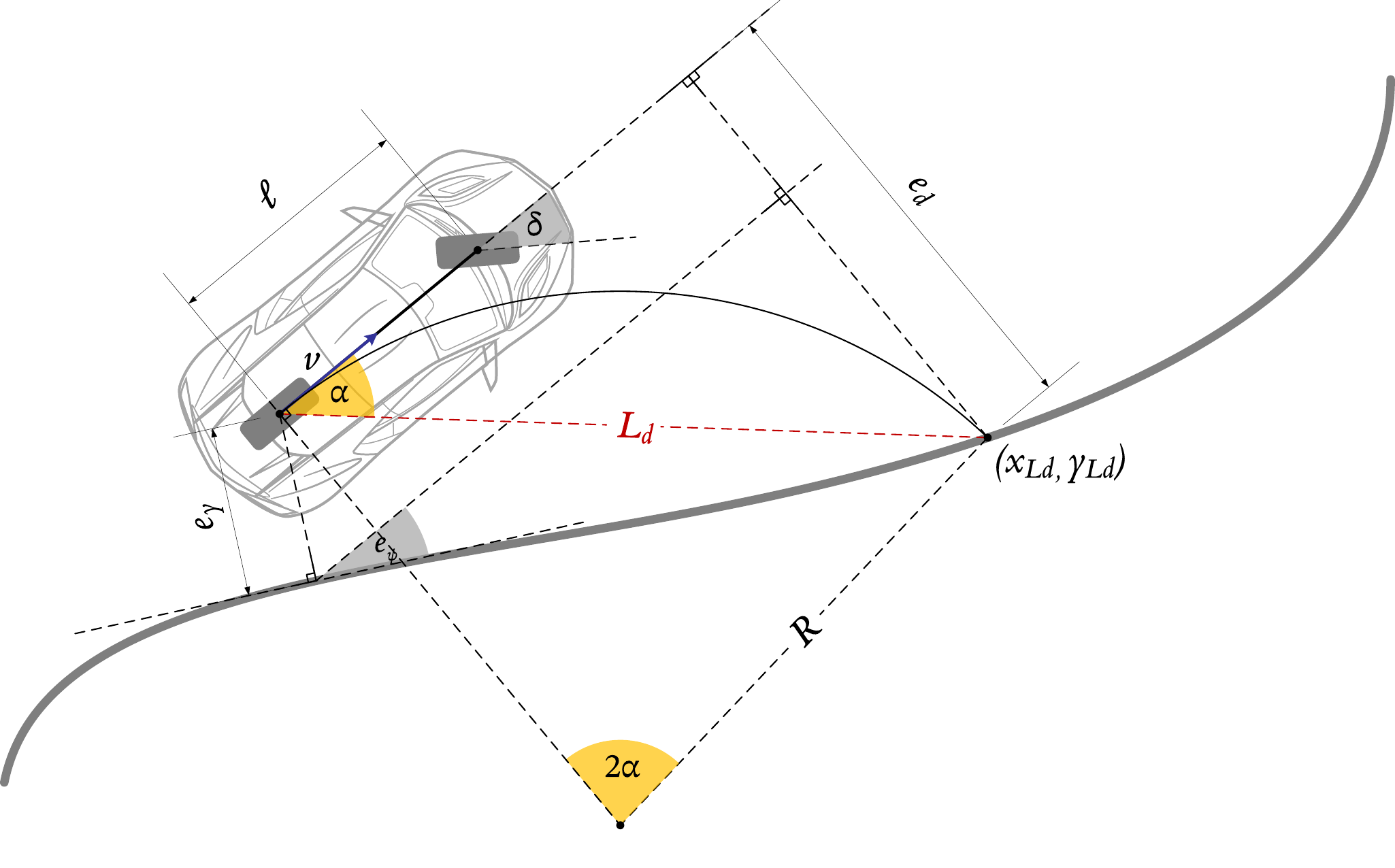}
	\caption{Pure Pursuit geometry}
	\label{fig:pp-geometry}
\end{figure}

Given the limitations of the reference scenario described above, the proposed VbLKS control architecture builds upon the well-established PP controller~\cite{Coulter1992}.
%
The PP control strategy can be derived starting from a bicycle model of the vehicle, by fitting an arc of circumference tangent to the vehicle's heading direction. The arc is defined as passing through the rear axis of the vehicle and a point on the reference path ahead of the vehicle by a specified distance $\lookahead$, known as the \textit{Lookahead distance}, as shown in Fig.~\ref{fig:pp-geometry}.

Based on this construction, a key quantity can be defined, namely the \emph{Lookahead Heading Error (LHE)} $\lhe$, which represents the angle between the car heading direction and the line joining the rear axis and the lookahead point on the path.
Alternatively, one can consider the \emph{Lookahead Lateral Error (LLE)} $\lle$, which represents the cross-track error between the point on the path and the heading direction of the vehicle.

\begin{figure}[!ht]
    \centering
    \includegraphics[width=1\linewidth]{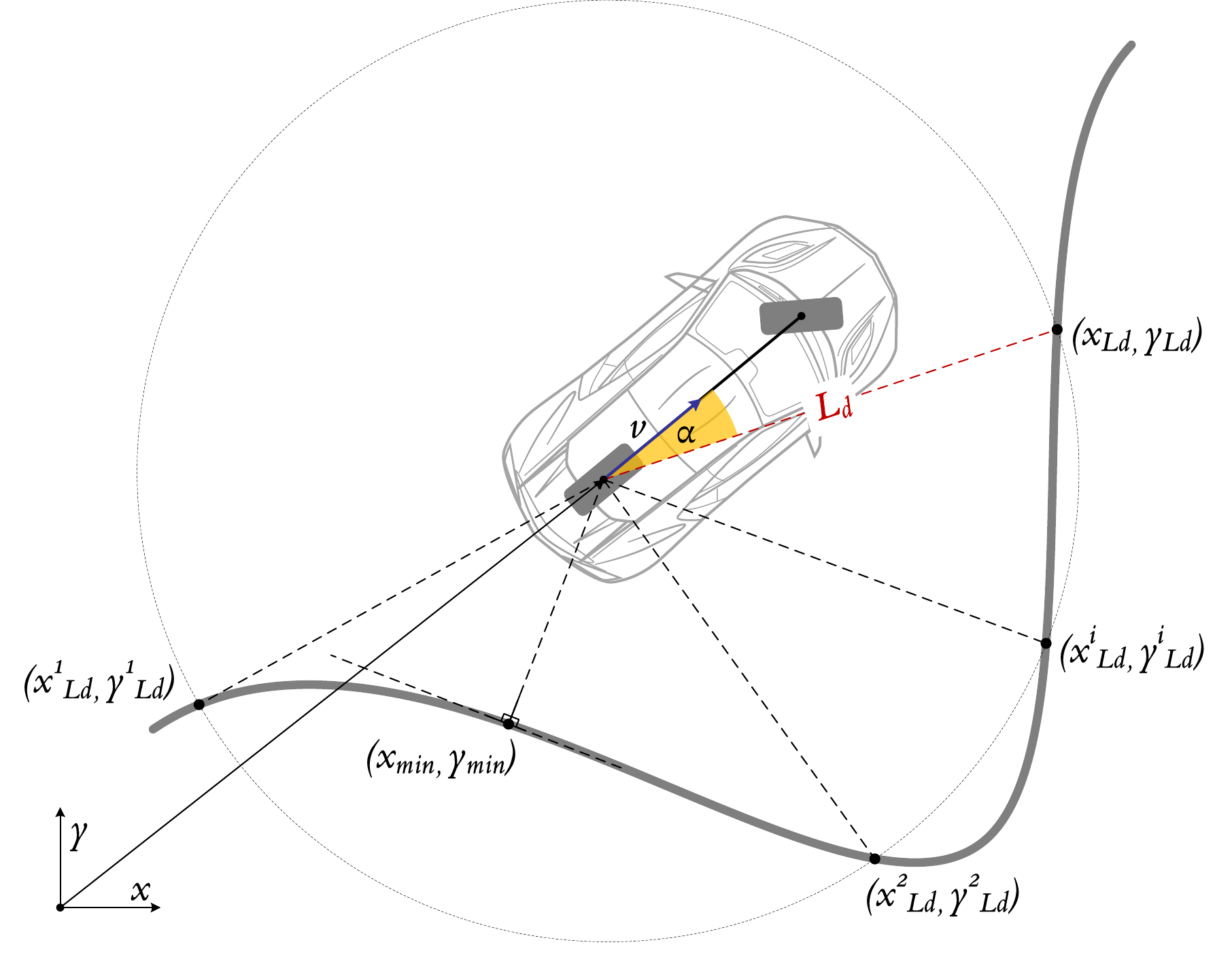}
    \caption{Heading error calculation}
    \label{fig:heading_error_calculation}
\end{figure}

Resorting to the rule of sines, the radius of the fitted arc can be derived as:
\begin{equation}\label{eq:pp-radius}
	R = \frac{1}{\curvature} = \frac{\lookahead}{2\sin\lhe},
\end{equation}
where $\curvature$ is the curvature of the interpolated arc of circumference.
Eventually, the Ackermann steering geometry assumption:
\begin{equation}\label{eq:Ackermann}
	\cot\steerangle_o - \cot\steerangle_i = \frac{\track}{\wheelbase},
\end{equation}
along with \eqref{eq:pp-radius} yields the PP control law:
\begin{equation}\label{eq:pp}
	\steerangle_r = \arctan\frac{2\wheelbase\sin\lhe}{\lookahead} = \arctan\frac{2\wheelbase\lle}{\lookahead^2}
\end{equation}
Assuming a global reference positioning system is available in Fig.~\ref{fig:pp-geometry}, another formulation of the LHE can be derived (see Fig.\ref{fig:heading_error_calculation}). Let's consider $\gamma$ to be the directed path parameterized on the curvilinear abscissa $s$ for the vehicle to follow, and $(x_r,y_r)$ the instantaneous position of the rear axis of the vehicle. Let $\mathcal{D}$ be the set of all the points on the path $(x_\lookahead^i, y_\lookahead^i)$ whose distance is $L_d$ meters from the rear axis:
\begin{equation}
     \mathcal{D} = \left\{ (x, y) \in \gamma \,\Big|\, (x - x_r)^2 + (x - y_r)^2 = \lookahead^2 \right\}.
\end{equation}
Under the assumption of $\left|\mathcal{D}\right| \geq 2$, i.e. there exist at least two points on the path with a distance of $\lookahead$ meters from the rear axis, the furthest point on the trajectory can be found as: 
    \begin{equation}
        \vect{p}_\lookahead = \argmax_{\vect{p}\in\mathcal{D}}{s},
    \end{equation}
and the global LHE $\lhe$ definition follows:
\begin{equation}
    \lhe = \atantwo{(y_\lookahead-y_r, x_\lookahead-x_r)} - \psi
\end{equation}
where $\psi$ is the heading angle of the vehicle.

In the proposed framework the LHE is estimated from camera frames through a small CNN (Fig.~\ref{fig:vblks-block-scheme}), which has been trained on a large amount of augmented visual data coming from the simulated environment.

The seamless transition of the computationally efficient CNN model from simulation to the real-world environment enables then the availability of a smooth real-time estimate of the LHE under experimental conditions, allowing for the introduction of a coupled path-tracking strategy, specifically tailored to improve tracking performance for increasing speed.






\section{CNN-based LHE Estimation}\label{sec:cnn-estimation}
The CNN-based estimation assumes a pivotal role in shaping the structure and novelty of the proposed methodology. As previously introduced, the primary objective of this layer involves the estimation of the LHE through the analysis of image data acquired from the frontal camera.
The input to the controller primarily relies on the estimated LHE, exerting a direct influence on the ultimate control performance. Consequently, our goal is to achieve high frequency (above the frame rate of the camera), low-latency, and accurate estimations.
A second consideration arises from the constrained computational power of the embedded device (Raspberry Pi 4B), coupled with the absence of specialized hardware for neural network inference. This compels the development of an efficient software pipeline, influencing the selection of neural network architecture, as well as the preprocessing and postprocessing steps, with an emphasis on mitigating computational demands.
Lastly, a fully Sim2Real training strategy is adopted, posing the challenge of generalizability to the real-world context but allowing for varied and comprehensive data collection.

The amalgamation of these three pivotal development decisions encapsulates the originality, innovation, and value of the proposed approach. These aspects manifest in three main procedural steps, to be elucidated in subsequent discussion: \textit{Data Collection}, \textit{Data Preprocessing and Augmentation}, \textit{Network Architecture and Training}.

\subsection{Data collection} \label{sec:data_collection}
The main idea of data collection is to use the simulation environment to effortlessly generate an arbitrarily large set of images and their corresponding ground truth labels. 
In order to reach this goal one must start from considering an appropriate path for the vehicle to follow, $\gamma$, which can be adapted depending on the scenario: in standard urban navigation the centerline of the lane can be used, while in racing scenario one could consider the optimal racing line. From a mathematical point of view $\gamma$ is defined as a directed path parametrized on the curvilinear abscissa $s$.
Starting from $\gamma$, a set of vehicle positions and orientations is generated by sampling a lateral and orientation error from two normal distributions. 
\begin{subequations}
\begin{IEEEeqnarray} {rCC}
    e_{iL} &=& \mathcal{N} (0, \sigma_L) \\ 
    e_{i\head} &=& \mathcal{N} (\head, \sigma_\head)\\  
    x_{Vi} &=& 0  
\end{IEEEeqnarray}
\end{subequations}

\begin{figure}
    \centering
    \makebox[\columnwidth][c]{\includegraphics[width=\columnwidth]{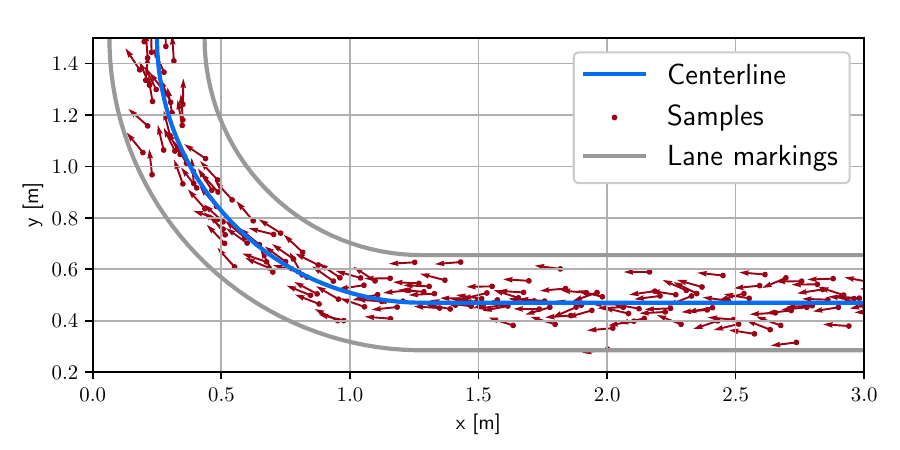}}
    \caption{Data sampling in simulation}
    \label{fig:original_vs_old}
\end{figure}

This is done in order to generate more variety in the training dataset. The noise injection is a key aspect of the estimation method and has a noticeable impact on the final performance, as it will be discussed in chapter \ref{sec:results}. The idea is that if the camera only sees the road ahead always perfectly centered, later, after the training, when facing the road at a different position and orientation, slightly off center, the network will not be able to estimate the LHE for lack of examples in the training dataset. The standard deviations of the noise injected will be subject of a parametric analysis. An example of the vehicle positions and orientations can be seen in Figure \ref{fig:original_vs_old}
The vehicle is moved around in the simulated environment by virtually moving it to the new desired position and orientation. For every position, an image from the virtual camera is saved alongside the vehicle position and orientation, in a compressed file for later use.
The ground truth LHE for every image is calculated using the method described in the previous section, which uses the saved positions and orientation and the lookahead distance to generate the LHE. 
This approach allows to generate an arbitrarily large number of samples and every frame can be used for multiple lookahead distances.

\subsection{Dataset Generation} \label{sec:dataset_gene}
This section delineates the procedure for generating the dataset employed in the training of the model. The dataset is comprised of pairs, encompassing input images and corresponding labels, which collectively serve as the training set for the model. The dataset generation involves two principal stages: \textit{Image Preprocessing} and \textit{Image Augmentation}. The overarching objectives of these operations are twofold: firstly, to alleviate the computational load and secondly, to enhance the model's capacity for generalization to real-world images, aligning with the Sim2Real paradigm.
    
\subsubsection*{Image Preprocessing} \label{sec:img_preprocessing}
\begin{figure}
    \centering
    \includegraphics[width=\linewidth]{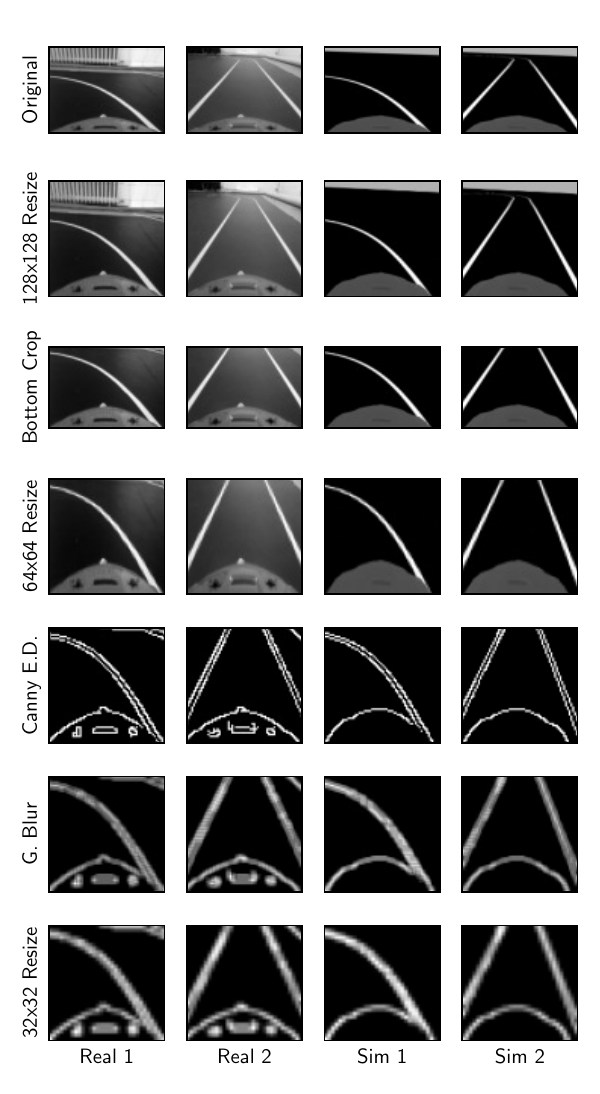}
    \caption{Sequence of preprocessing steps applied to a pair of images from the camera and their simulated counterparts.}
    \label{fig:preprocessing_steps}
\end{figure}

Image preprocessing is applied both to the images collected in the simulator and to the images processed in real time for online estimation. The main goals of preprocessing are to reduce the amount of information in the image to the minimum required for the task, standardize the images to make the real and simulated data more comparable, and reduce computational cost. The sequence of the preprocessing steps (Figure \ref{fig:preprocessing_steps}) is the following: gray scale conversion, bottom crop, resize to 64x64, Canny edge detection \cite{canny}, Gaussian blurring, resize to 32x32. The key step in term of reducing the Sim2Real gap is played by the Canny edge detection, that drastically reduces the difference between real and simulated images by removing the light information.
    
\subsubsection*{Image augmentation}
\begin{figure}
    \centering
    \includegraphics[width=\linewidth]{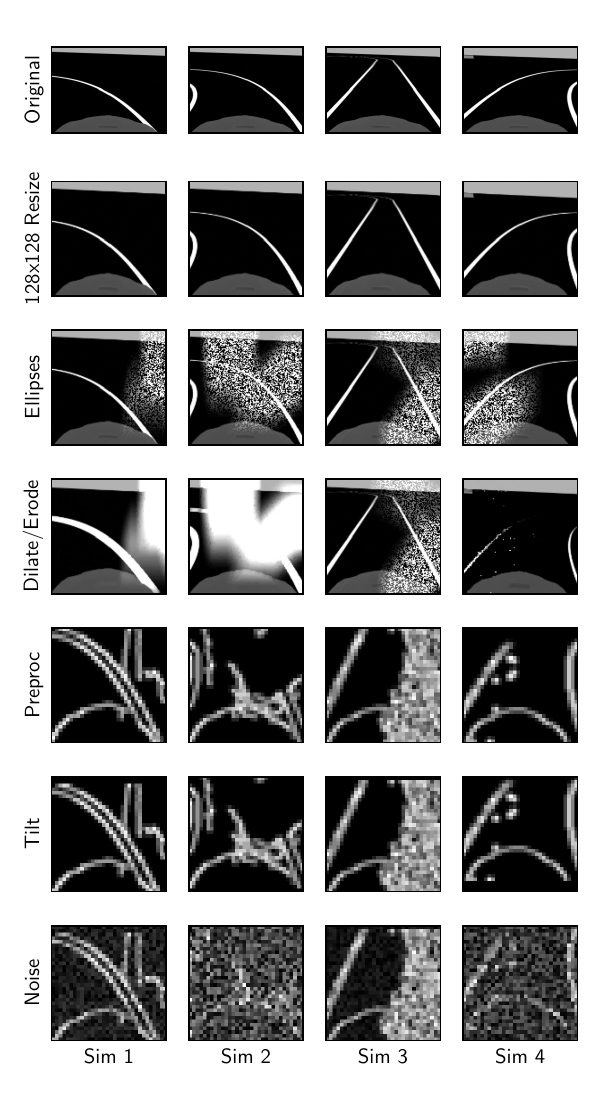}
    \caption{Sequence of augmentation steps applied to 4 images from the simulator.}
    \label{fig:augmentation_steps}
\end{figure}

Image augmentation is a technique used to artificially increase the size of the dataset by slightly altering the images in the dataset. The main goals of image augmentation are to increase the number of images and make the simulator images more similar to the real ones.
We observe that the problem of estimating the LHE from a frontal image is symmetric with respect to the vertical axis. This means that we can flip the image horizontally and invert the sign of the LHE to obtain a new valid sample. 
In addition to this symmetry, the data augmentation process consists in a series of steps showed in Figure \ref{fig:augmentation_steps}:
    
\begin{enumerate}
    \item \textbf{Resize}: The images are resized to 4 times their final size, resulting in a 128x128 square.
    \item \textbf{Random ellipses}: Random ellipses are drawn on the image with random positions, orientations, and shades of white. This simulates strong light reflections, which are not present in the simulated environment. The ellipses also partially cover the image, serving as a form of random erasing \citep{random_erase}, a common data augmentation technique used to reduce overfitting and increase robustness to occlusions.
    \item \textbf{Dilation/erosion}: The images are dilated or eroded (with a 20\% probability for each) using a random kernel.
    \item \textbf{Preprocessing}: The images are preprocessed as described in the previous section.
    \item \textbf{Random shift}: The images are randomly shifted by a small number of pixels in the vertical direction.
    \item \textbf{Noise}: The images are randomly corrupted with Gaussian noise.
\end{enumerate}

\subsection{Neural Network design and training} \label{sec:nn_training}
In this section, we elucidate the design of the neural network and the training process employed. It is imperative to formulate the network with the constraint that it must execute in real-time on an embedded system. Therefore, the network must be compact while still exhibiting high performance and generalization capabilities from simulated data to real data.
\subsubsection*{Architecture}
To fulfill the real-time performance requirements, the neural network is crafted with a minimal number of layers and parameters. The architecture draws inspiration from LeNet-5 \citep{lecun}, a straightforward convolutional network initially devised for handwritten digit recognition. The design is enhanced by the incorporation of batch normalization \citep{batch_norm} and dropout \citep{dropout}. The network comprises a sequence of convolutional layers succeeded by a sequence of fully connected layers. Input to the network consists of 32x32 grayscale images, and the output is a single floating-point value representing the LHE at a fixed distance, defined during training.
\subsubsection*{Training}
The training process adheres to conventional standards, employing an 80/20 train-validation split. It is essential to note that, in alignment with the Sim2Real paradigm, no data from the real world has been utilized for training or validation. Real-world data is exclusively employed for evaluating Sim2real and final performance.
Both the architecture and training details are succinctly summarized in Table \ref{tab:nn_architecture}.

\begin{table}[h]
\centering
\begin{tabular}{p{0.3\linewidth}p{0.6\linewidth}}
    \hline
    \textbf{Architecture} &  \\
    \hline
    \textbf{1st 2D convolution} & 4 filters, with a kernel size of 5x5 and a stride of 1. With ReLu activation function. \\
    \hline
    \textbf{Dropout} & 30\% probability. \\
    \hline
    \textbf{Pooling} & Max pooling with a kernel size of 2x2 and a stride of 2. \\
    \hline
    \textbf{Batch normalization} & Normalization of the output of the previous layer. \\
    \hline
    \textbf{2nd 2D convolution} & 16 filters, with a kernel size of 5x5 and a stride of 1. With ReLu activation function. \\
    \hline
    \textbf{Dropout} & 30\% probability. \\
    \hline
    \textbf{Pooling} & Max pooling with a kernel size of 2x2 and a stride of 2. \\
    \hline
    \textbf{Dropout} & 30\% probability. \\
    \hline
    \textbf{3rd 2D convolution} & 32 filters, with a kernel size of 5x5 and a stride of 1. With ReLu activation function. \\
    \hline
    \textbf{Flatten} & Flattens the input into a vector of 32 elements. \\
    \hline
    \textbf{1st fully connected} & 16 neurons. With ReLu activation function. \\
    \hline
    \textbf{2nd fully connected} & Output layer, 1 neuron. Without any activation function. \\
    \hline
    \hline
    \textbf{Training} &  \\
    \hline
    \textbf{Dataset} & 80 \% train/validation split \\
    \hline
    \textbf{Loss function} & Mean squared error (MSE)\\
    \hline
    \textbf{Optimizer} & Adam optimizer, learning rate of $\num{3e-3}$. \\
    \hline
    \textbf{Duration} & 200 epochs, batch size of $2^{16}$. \\
    \hline
    
\end{tabular}
\caption{Neural Network Architecture and Training}
\label{tab:nn_architecture}
\end{table}

\subsubsection*{Inference}
In this section, the process of inference on the embedded device, involving the execution of the neural network on the real car hardware, is elucidated. The PyTorch model is converted to an ONNX model, enabling direct utilization with the OpenCV library and alleviating the need to install the resource-intensive PyTorch library on the embedded device.
For online estimation, the camera image undergoes preprocessing and is inputted to the network, which subsequently predicts the LHE. To enhance estimation accuracy, the image is horizontally flipped, and the network is executed twice, once with the original image and once with the flipped image. The average of the resulting measurements is computed, accounting for the symmetry inherent in the problem.
In terms of performance, the network achieves a frame rate of 250-300 FPS on the Raspberry Pi 4, exceeding the requirements for real-time estimation. The limiting factor, in this case, is the frame rate of the camera. The efficiency of the architecture suggests suitability for deployment on even lower-end devices.


\section{PP-based Control}\label{sec:pp-control}
In this section we delve deeper into the path tracking control strategy based on the LHE estimate coming from the CNN at a constant lookahead distance, which has been described in the previous section. 
As introduced in Sec.\ref{subsec:geomform}, the VbLKS is founded on the Pure Pursuit strategy, whose main shortcoming resides in selecting the lookahead distance. Too short $ \lookahead $ enhances accuracy in following the path, but may trigger oscillatory behaviours, while too large $\lookahead$ leads to smoother tracking but may result in the undesirable ``corner-cutting" effect as the vehicle approaches tight curves \cite{Snider2009}. Indeed, the underlying geometric assumptions of this controller disregard any variations in the curvature of the path ahead, considering it to be constant.
A popular strategy for balancing stability and tracking performance involves adjusting the lookahead distance according to the velocity 
\cite{Snider2009}. For instance, in a recent study by \citeauthor{Sukhil2021} \cite{Sukhil2021}, they employed a labelling policy to allocate varying lookahead distances to specific sections of the reference track, based on desired racing objectives.

In our case, the lookahead distance in constrained to be fixed by the estimation part.
Therefore, in the subsequent sections, we describe the PP-D and PP-VR control strategies, which are designed to address both this constraint and the delay inherent in the steering actuation system with the focus of enhancing tracking performance for increasing speed. Stability analysis is used for developing a tuning strategy for the controller parameters.


\subsection{Pure Pursuit Derivative controller (PP-D)}
Resorting to the first order Taylor approximation of the $ \arctan $ and $ \sin $ functions, the PP control law \eqref{eq:pp}  can be linearized as:
\begin{equation}\label{eq:pp-linear}
	\steeranglelin_r = \frac{2\wheelbase}{\lookahead}\lhe
\end{equation}
Under small LHE assumption, it holds true that $ \lhe\approx \dotlaterr/\vel + \laterr/\lookahead $ so that the linearized PP controller is defined as:
\begin{equation}\label{eq:pp-pd-lin}
	\steeranglelin_r = \frac{2\wheelbase}{\vel\lookahead}\left(\dotlaterr+\frac{\vel}{\lookahead}\laterr\right)
	\quad \underset{\mathcal{L}}{\longrightarrow}\quad 
	\frac{2\wheelbase}{\lookahead^2}\,\left(1 + s\frac{\lookahead}{\vel}\right)\,E_y(s)
\end{equation}
From this point of view the gain, which is inversely proportional to the square of $\lookahead$, governs the aggressiveness of the control action; on the other hand, the stable zero depends on the ratio of $\lookahead$ to the velocity and is responsible for regulating the anticipative action, thus enabling phase recovery around the natural frequency of the system \cite{Park04}.
To cope with the constraint of fixed lookahead distance, a derivative action on the LHE can be introduced, thanks to a low-noise input provided by the CNN. This results in the PP-D control law:
\begin{equation}\label{eq:pp-pd}
	\begin{IEEEeqnarraybox}[][t]{C}
		\steerangle_{r,d} = \arctan\frac{2\wheelbase\sin\lhe}{\lookahead} + \ppgainder\frac{\diff{\lhe}}{\dt}\quad, \quad \ppgainder>0
	\end{IEEEeqnarraybox}
\end{equation}
By linearizing \eqref{eq:pp-pd}, it can be observed that the derivative action introduces an additional stable zero, which enables it to influence the anticipatory behaviour of the controller:
\begin{equation}\label{eq:pp-pd-linearized-tf}
	\steeranglelin_{r,d} = \frac{2\wheelbase}{\lookahead^2}\left(1+s\frac{\lookahead}{\vel}\right) \left(1+s\frac{\lookahead}{2\wheelbase}\ppgainder\right)\,E_y(s)
\end{equation}
In the next sections we will show how the added derivative action not only plays a crucial role in forecasting the future dynamics of the LHE but also helps in preventing oscillations caused by the actuation delay.
However, it is essential to recognize that adjusting the $\ppgainder$ gain is not equivalent to tuning the lookahead distance. Indeed, the lookahead distance contains additional information about the path ahead and continues to play a vital role in the tuning process.

\subsection{Vehicle model for PP stability analysis}\label{sec:vehicle-model}
\begin{figure}[!ht]
	\centering
	\subfloat[Global]{\includegraphics[width=.4\columnwidth]{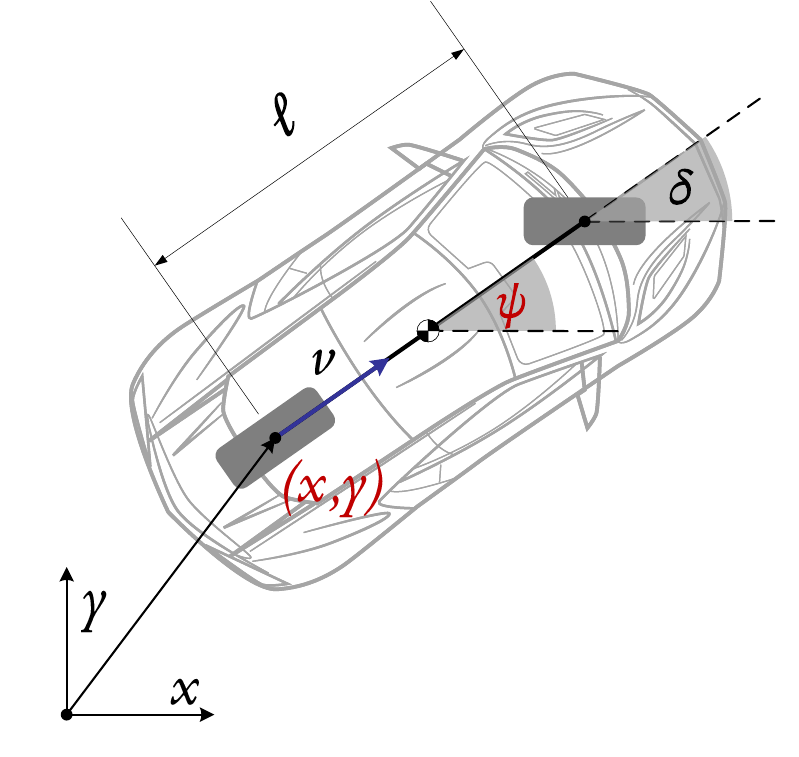}
		\label{fig:bicycle-global}}
	\hfil
	\subfloat[Local]{\includegraphics[width=.5\columnwidth]{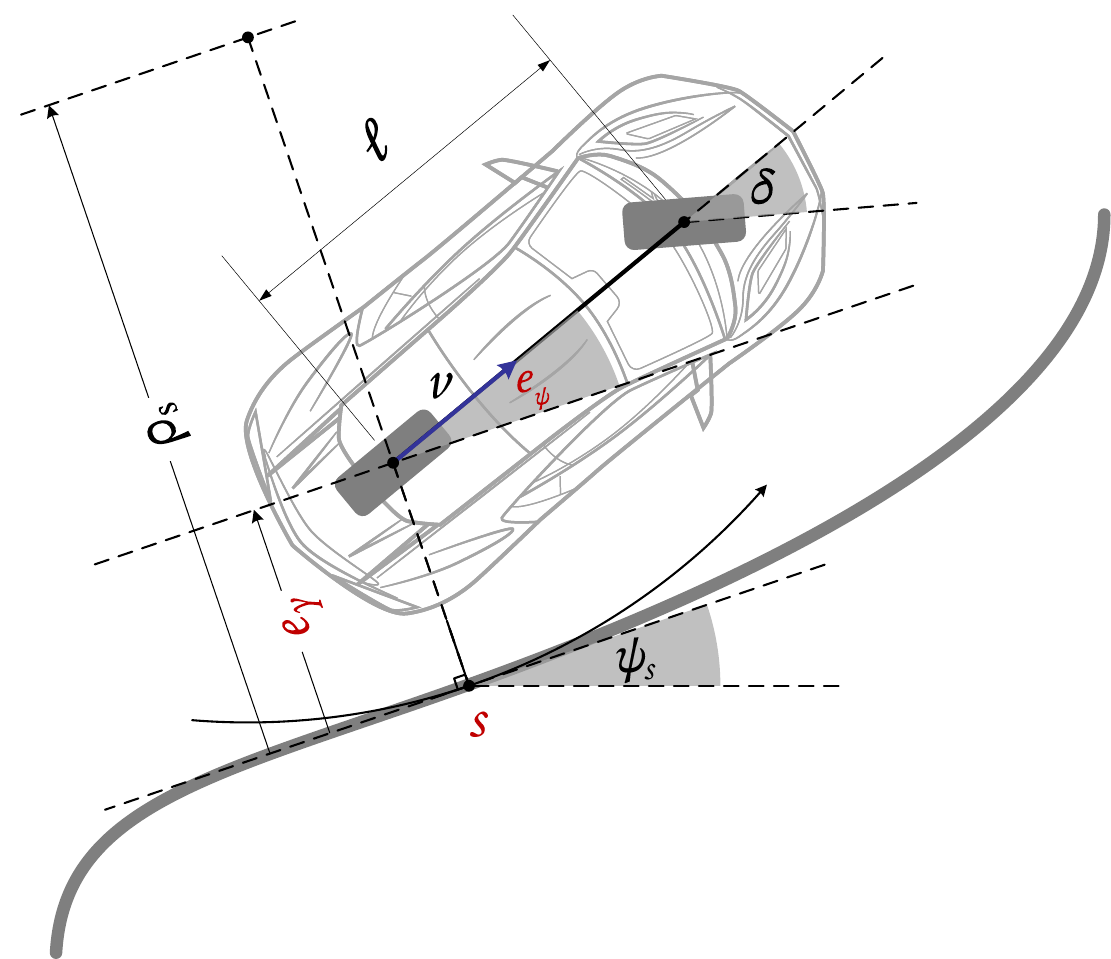}
		\label{fig:bicycle-local}}
	\caption{Bicycle model of the vehicle}
\end{figure}

In this section, the model of the vehicle used in the delay stability analysis is described. In view of a PP type of control, a kinematic model of the vehicle is used in this article.
The equations that describe a driftless, rear wheel driving bicycle model are \cite{DeLuca1998}:
\begin{subequations}\label{eq:bicycle-model-global}
	\begin{IEEEeqnarray}{rCCCl}
		\dot{x}(t) &=& \frac{\diff{x(t)}}{\dt} &=& \vel\cos\head(t)\\
		\dot{y}(t) &=& \frac{\diff{y(t)}}{\dt} &=& \vel\sin\head(t)\\
		\dot{\head}(t) &=& \frac{\diff{\head(t)}}{\dt} &=& \frac{\vel}{\wheelbase}\tan\head(t)
	\end{IEEEeqnarray}
\end{subequations}
where $ \left[x,y,\head\right]^\top $ represents the pose of the vehicle with respect to a world reference frame (Fig.\ref{fig:bicycle-global}).

In the path tracking scenario, it is useful to consider the relative pose of the vehicle with respect to a given path, therefore an error model in path coordinates can be derived from \eqref{eq:bicycle-model-global}:
\begin{subequations}\label{eq:bicycle-model-local}
	\begin{IEEEeqnarray}{rCl}
		\dot{s} &=& \frac{\pathradius(s) \vel\cos\headerr}{\pathradius(s)-\laterr}\\
		\dotlaterr &=& \vel\sin\headerr\\
		\dotheaderr &=& \frac{\vel}{\wheelbase}\tan\steerangle - \frac{v\cos\headerr}{\pathradius(s)-\laterr} \eqdef \frac{\vel}{\wheelbase}\tan\steerangle - \dot{\head}_s(s)
	\end{IEEEeqnarray}
\end{subequations}
where $ s $ is the projected vehicle position along the lane center line, $ \pathradius $ is the radius of curvature of the road and $ \head_s $ is the road heading angle.
The nonlinear local model \eqref{eq:bicycle-model-local} can be recast from a control point of view as:
\begin{subequations}\label{eq:model-pp-control}
	\begin{IEEEeqnarray}{rCl}
		\dotx &\defeq
        & \begin{bmatrix}
            \laterr \\ \headerr
        \end{bmatrix}
        = \begin{bmatrix}
			\vel\sin\headerr\\
			\dfrac{\vel}{\wheelbase}\tan\innput-\dot{\head}_s(s)
		\end{bmatrix}\\
		\outtput &=& \arcsin\frac{\lle(\state)}{\lookahead}
	\end{IEEEeqnarray}
\end{subequations}
where the steering angle $ \steerangle $ and the LHE $ \lhe $ are respectively the input and the output of the system. In general, the lack of a closed-form characterization of the output function prevents a comprehensive study of the closed-loop control system. However, based on \eqref{eq:model-pp-control}, a favourable scenario can be derived.

\begin{figure}[!ht]
	\centering
	\includegraphics[width=.9\columnwidth]{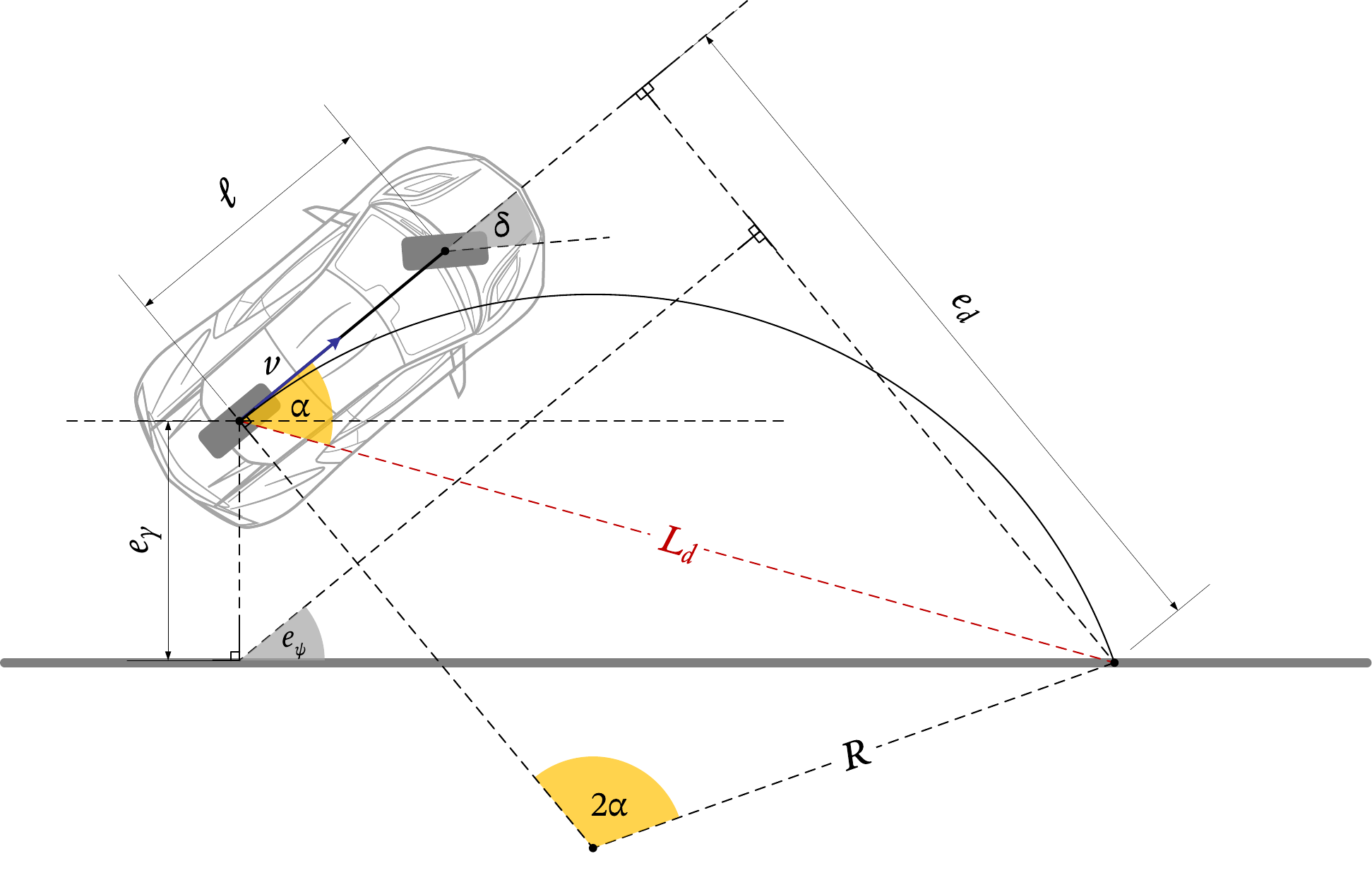}
	\caption{Straight scenario}
	\label{fig:pp-straight}
\end{figure}

In the case of a straight path (Fig.\ref{fig:pp-straight}), the LLE can be derived by resorting to geometric rules. The LLE takes the form:
\begin{equation}\label{eq:lle-straight}
	\lle = \laterr\cos\headerr + \sqrt{\lookahead^2-\laterr^2} \sin\headerr.
\end{equation}

In this symmetric scenario, $ \head_s $ is zero and \eqref{eq:model-pp-control} simplifies to:
\begin{subequations}\label{eq:model-pp-control-straight}
	\begin{IEEEeqnarray}{rCl}
		\dotx &=& \begin{bmatrix}
			\vel\sin\headerr\\
			\dfrac{\vel}{\wheelbase}\tan\steerangle
		\end{bmatrix}\\
		\outtput &=& \arcsin\frac{\laterr\cos\headerr + \sqrt{\lookahead^2-\laterr^2} \sin\headerr}{\lookahead}
	\end{IEEEeqnarray}
\end{subequations}
The origin $ [0,0]^\top $ with both null input and output form an equilibrium configuration for \eqref{eq:model-pp-control-straight} which can be linearized thus yielding the transfer function model:
\begin{equation}\label{eq:fdt-straight}
    P_S(s) = \frac{v^2}{\wheelbase\lookahead}\frac{1+s\frac{\lookahead}{v}}{s^2}
\end{equation}

The model for control \eqref{eq:model-pp-control} lacks a comprehensive description of the steering system, as it assumes that the input steering signal instantaneously affects the system. To address this limitation, a first-order ODE can be used to incorporate both the static delay and dynamics of the steering mechanism \cite{Xu2021}:
\begin{equation}
	\steerlag\dot{\steerangle}(t) = -\steerangle(t-\steerdelay) + \steerangle_r(t)
\end{equation}
In this model, the \emph{static delay} $ \steerdelay $ represents the time in between when the reference signal is applied and when the steering wheel starts turning; whereas the dynamics the wheel shows once it starts turning, can be modelled by a first-order system with a time constant $ \steerlag $, hereinafter referred to as \emph{steering lag}. The corresponding first order plus time delay transfer function from the steering angle reference to the actual steering angle can be derived as:
\begin{equation}
	D(s) = \frac{e^{-s\steerdelay}}{1+s\steerlag}
\end{equation}

\subsection{Stability analysis in the straight scenario and tuning}\label{sec:delay-analysis}
This section provides a time-delay stability analysis in the straight scenario, with a particular focus on speed performance.
To this end, the Walton-Marshall (WM) direct method \cite{Walton1987} can be employed as an approximate tool for numerical analysis. This method assists in determining the values of static steering delay that result in a stable system, considering fixed control and system parameters. The WM analysis is further employed to establish tuning bounds, providing guidance for implementing the controller on the real car.
Studying the stability of delay-free systems is relatively straightforward since the number of roots in their characteristic equations is finite. However, the introduction of time delays results in an infinite number of roots, making the task of establishing stability significantly more challenging.
To address this challenge, the Hermite-Bieler Theorem for Quasi Polynomials offers necessary and sufficient conditions for determining whether the roots of the characteristic equation with time delay have a negative real part \cite{Silva2005}.
In the specific context of time-delay systems with a single delay, Walton and Marshall proposed an alternative analysis in \cite{Walton1987}. Such systems are characterized by the following characteristic equation:
\begin{equation}\label{eq:characteristic-time-delay}
	\Delta(s) = d(s) + n(s)e^{-s\steerdelay}
\end{equation}
where $ d(s) $ and $ n(s) $ are coprime polynomials with real coefficients, $ \deg{d(s)}=q $, $ \deg{n(s)}=p $ and $ q\geq p $.

This method can be divided into three main steps:
\begin{enumerate}
	\item Examine the stability of \eqref{eq:characteristic-time-delay} for $ \steerdelay=0 $ using classical methods
	\item Consider the case of infinitesimally small $ \steerdelay $: the number of roots changes from being finite to infinite. If $ q>p $, all new roots of \eqref{eq:characteristic-time-delay} must lie in the Left Half Plane (LHP) and this step can be omitted. If $ q=p $, more details are involved \cite{Nguyen2019}.
	\item Find $ \steerdelay>0 $, if any, at which there are roots lying on the imaginary axis and determine whether these roots merely touch the axis or if they cross from one HP to the other for increasing $ \steerdelay $. To do so, determine the positive roots $ \w^\star $ of
	\begin{equation}
		Q(\w^2) \defeq d(j\w)d(-j\w) - n(j\w)n(-j\w),
	\end{equation}
	determine the corresponding positive values of $ \steerdelay $ s.t. $ e^{-j\w^\star\steerdelay} = -d(j\w^\star)/n(j\w^\star) $ and the nature of these roots by inspecting
	\begin{equation}
		S=\sign{\left[\left.\frac{\diff Q(\w^2)}{\diff \w^2}\right|_{\w=\w^\star}\right]}=
		\begin{cases}
			+1 & \w^\star\text{ destabilizing}\\
			-1 & \w^\star\text{ stabilizing}
		\end{cases}
	\end{equation}
\end{enumerate}
Applying the PP-D derivative control law \eqref{eq:pp-pd} to the transfer function model in the straight scenario \eqref{eq:fdt-straight}, the characteristic equation of the linearized feedback system becomes:
\begin{equation*}
    \begin{aligned}
        \Delta(s) &\overset{\hphantom{\steerdelay=0}}{\defeq} d(s)+n(s)e^{-s\steerdelay}\\
    	&\overset{\hphantom{\steerdelay=0}}{=} s^2\left(1+s\steerlag\right) + \frac{v^2}{\wheelbase\lookahead}\left(\frac{2\wheelbase}{\lookahead} + s\ppgainder\right)\left(1+s\frac{\lookahead}{v}\right)e^{-s\steerdelay}\\
    	&\overset{\steerdelay=0}{=} \steerlag\,s^3 + \left(1+\frac{v}{\wheelbase}\ppgainder\right)\,s^2
    	+ \left(\frac{2v}{\lookahead} + \frac{v^2}{\wheelbase\lookahead}\ppgainder\right)\,s + \frac{2v^2}{\lookahead^2}
    \end{aligned}
\end{equation*}
Resorting to the Routh table in the delay free case, the Routh-Hurwitz criterion yields the following conditions:
\begin{equation*}
	\begin{cases}
		\lookahead > \underbrace{\left(\frac{2}{\left(2+\ppgainderPD\right)\left(1+\ppgainderPD\right)}\right)}_{\mu(\ppgainder)}\,v\steerlag\\
		\ppgainprop > 0
	\end{cases}
\end{equation*}
where the second condition comes from the assumption of a stable controller, and for sake of clarity, we defined $ \ppgainderPD=\ppgainder\frac{v}{\wheelbase} $.
The constraint on the lookahead distance found by \citeauthor{Ollero1995} in \cite{Ollero1995} is here integrated by the function $ \mu(\ppgainder) $ which is upper bounded by 1 in the null derivative case.
As the derivative gain increases, for fixed proportional gain, the bound on $ \lookahead $ is relaxed, thus allowing for increasing the speed, for fixed lookahead distance.
Given the substitution $ \wsq = \w^2 $:
\begin{equation}
	\begin{aligned}
		Q(\wsq) &\defeq d(\jw)d(-\jw) - n(\jw)n(-\jw)\\
		&= \steerlag^2\,\wsq^3 + 
		\left(1-\ppgainder^2\frac{v^2}{\wheelbase^2}\right)\,\wsq^2 - \cdots\\
		&\hphantom{=} \quad \cdots \left(\frac{4v^2}{\lookahead^2}+\ppgainder^2\frac{v^4}{\wheelbase^2\lookahead^2}\right)\,\wsq - \frac{4v^4}{\lookahead^4}
	\end{aligned}
    \label{eq:WM-poly-straight}
\end{equation}
The WM polynomial $ Q(\wsq) $ has one degree of freedom in the derivative gain.
The derivative of \eqref{eq:WM-poly-straight} with respect to $\wsq$ follows as:
\begin{equation}
    \begin{aligned}
        \frac{\diff{Q}}{\diff{\wsq}} &=
        3\steerlag^2\,\wsq^2 + 
        2\left(1-\ppgainder\frac{v^2}{\wheelbase^2}\right)\,\wsq - \cdots \\
        &\hphantom{=} \quad \cdots \left(\frac{4v^2}{\lookahead^2}+\ppgainder^2\frac{v^4}{\wheelbase^2\lookahead^2}\right)
    \end{aligned}
    \label{eq:WM-poly-straight-der}
\end{equation}
Evaluating the roots of \eqref{eq:WM-poly-straight} and the sign of \eqref{eq:WM-poly-straight-der} as a function of the controller parameters allows then to determine the critical delay of the system, if it exists, up to which of the linearized closed loop system in the straight scenario is stable.

According to \eqref{eq:pp-pd-linearized-tf} the anticipative action of the PP-D controller depends also on velocity. Consequently, in the following Section, a strategy for generating a variable velocity reference is presented. 

\subsection{Pure Pursuit Velocity Reference (PP-VR)}
In order to guarantee homogeneity along motion and safer transitions between straight and curved sections, a technique inspired by \cite{Filho2014} is used for longitudinal control. The aim of this technique is that of driving the vehicle at a desired speed $ \velmax $, avoiding excessive lateral acceleration on corners. The control law proposed in \cite{Filho2014} is:
\begin{equation}\label{eq:speed-ref-article}
	v = \min\left\{\velmax,\sqrt{\frac{\lataccmax}{\kappa_s(t)}}\right\}
\end{equation}
where $ \kappa_s(t) $ is the curvature of the road and $ \lataccmax $ is the maximum lateral acceleration of the vehicle. In this work, lateral acceleration is computed so that the vehicle's velocity converges exponentially to the desired goal velocity. This approach requires knowing the current curvature of the path which is not available in our scenario. 
However, Ackermann geometry \eqref{eq:Ackermann} assumes that the curvature between the current position and the lookahead point is considered to be constant along motion, with radius \eqref{eq:pp-radius}. 
The relation between $ \lhe $ and $ \latacc $ allows then to extend \eqref{eq:speed-ref-article} to:
\begin{equation}\label{eq:pp-vr}
	v = \min\left\{\velmax,\sqrt{\frac{\lookahead}{2\sin\lhe}\lataccmax}\right\}.
\end{equation}
For give $\velmax$ and $\lataccmax$ as control parameters, this law can be used as a variable velocity reference to an inner loop based on a PID controller with feedback from an encoder mounted on the motor shaft.

\section{Experiments and Results} \label{sec:results}
This section outlines the experimental framework and presents the outcomes of the proposed VbLKS. 
The following key aspects will be discussed:
\begin{itemize}
    \item \textbf{Experimental setup}: Details regarding the real track and the localization system employed for data collection and system testing.
    \item \textbf{PP-based Controller Tuning}: a tuning strategy for the PP-based controller parameters.
    \item \textbf{CNN Hyperparameters Tuning and Performance Evaluation}: validation of the CNN by means of simulated and real test datasets. The effectiveness of the Sim2Real approach is validated as well, comparing optimal hyperparameters tuning on simulated and real datasets.
    \item \textbf{Performance on test track}: Discussion of experiments conducted with the full VbLKS on the test track.
\end{itemize}

\subsection{Experimental setup: Real track and data collection with Vicon localization system} \label{sec:exp_setup}

In order to precisely determine the location and orientation of the vehicle, a Vicon system is employed. This system encompasses an array of cameras strategically positioned to track any object equipped with markers. Notably, this system, characterized by a frequency of 100 Hz and minimal packet loss, facilitates seamless integration into an existing ROS network.

\begin{figure}
    \centering
    \includegraphics[width=\linewidth]{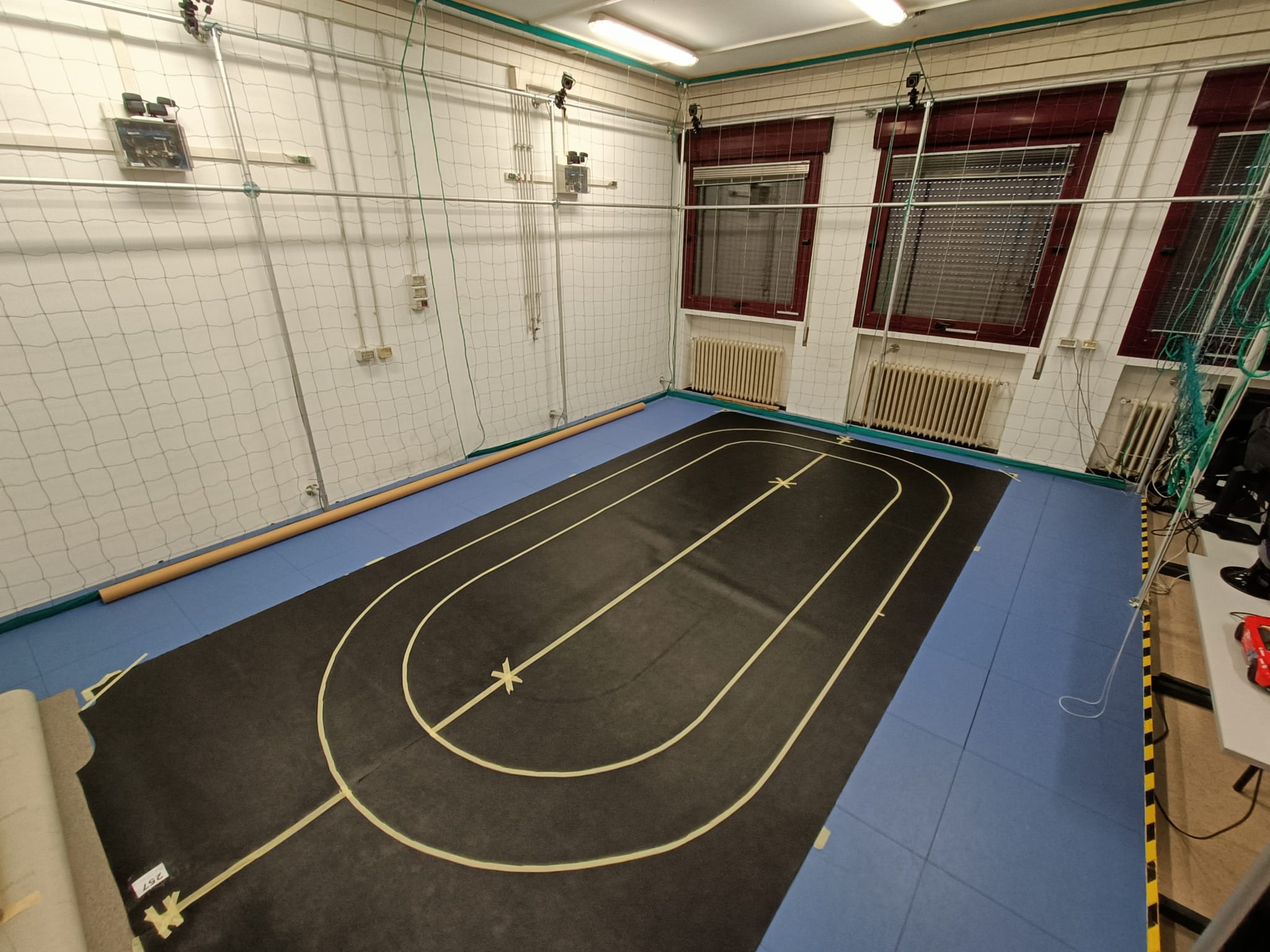}
    \caption{Test track}
    \label{fig:track}
\end{figure}

To test the vehicle, we designed a track consistent with BFMC specifics, consisting of three straight segments connected by three turns, as shown in Figure \ref{fig:track}. The track is about 12 meters in length, with longer straight segments measuring 2 meters. The track surface is black, with white tape lines that are 2 cm wide and spaced 37 cm apart marking the path. The same track was reproduced in a simulated environment, enabling us to compare the real and simulated data obtained by driving the same trajectories.
The centerline of the track can be parameterized by the following equations:

\begin{small}
\begin{align*}
    \vect{r}_{12}(\theta)
    &=
    \begin{bmatrix}
        x_c + R\cos\theta\\
        \bigl(m+R_c\bigr)+R\sin\theta
    \end{bmatrix}
    & \theta\in[0,-\pi]\\
    \vect{r}_{23}(\lambda)&=
    \begin{bmatrix}
        x_c - R\\
        \lambda\bigl(m+R_c\bigr)+(1-\lambda)\bigl(m+R_c+L\bigr)
    \end{bmatrix}
    & \lambda\in[0,1]\\
    \vect{r}_{34}(\theta)&=
    \begin{bmatrix}
        \bigl(x_c - R + r\bigr) + r\cos\theta\\
        \bigl(m+R_c+L\bigr) + r\sin\theta
    \end{bmatrix}
    & \theta\in\left[-\pi,-\frac{3\pi}{2}\right]\\
    \vect{r}_{45}(\lambda)&=
    \begin{bmatrix}
        \lambda\bigl(x_c-R+r\bigr)+(1-\lambda)\bigl(x_c+R-r\bigr)\\
        m+R_c+L+r
    \end{bmatrix}
    & \lambda\in[0,1]\\
    \vect{r}_{56}(\theta)&=
    \begin{bmatrix}
        \bigl(x_c + R - r\bigr) + r\cos\theta\\
        \bigl(m+R_c+L\bigr) + r\sin\theta
    \end{bmatrix}
    & \theta\in\left[-\frac{3\pi}{2},-2\pi\right]\\
    \vect{r}_{61}(\lambda)&=
    \begin{bmatrix}
        x_c + R\\
        \lambda\bigl(m+R_c+L\bigr)+(1-\lambda)\bigl(m+R_c\bigr)
    \end{bmatrix}
    & \lambda\in[0,1]\\
\end{align*}
\end{small}

With $\vect{r}_{ij}(\cdot)$ representing each section of track, and with the symbols descriptions
and numerical values summarized in Table \ref{tab:track_parameters}.
\begin{table}[!ht]
    \centering
    \begin{tabular}{>{$}c<{$}@{\hspace{1em}}l>{$}c<{$}}
        \toprule
        \textbf{Sym} & \textbf{Description} & \textbf{Value}\\\toprule
        x_c 	& x coordinate of the axis of symmetry               & 1.5\,\si{\meter}\\\midrule
        m 		& Margin from the edge of the fabric	                        & 0.25\,\si{\meter}\\\midrule
        w		& Lane width						& 0.37\,\si{\meter}\\\midrule
        L		& Length of the straight section					& 2\,\si{\meter}\\\midrule
        R_c 	& Big radius of the center of the lane				& 1.04\,\si{\meter}\\\midrule
        R_i,R_e 		& Internal and external big radii & \left\{R_c-\frac{w}{2},R_c+\frac{w}{2}\right\}\\\midrule
        r_c		& Small radius of the center of the lane			& 0.65\,\si{\meter}\\\midrule
        r		& Internal and external small radii 							& \left\{r_c-\frac{w}{2},r_c+\frac{w}{2}\right\}\\\bottomrule
    \end{tabular} 
    \caption{Parameters of the test track} \label{tab:track_parameters}
\end{table} 

\subsection{PP-based Controller Tuning}\label{subsec:tuning-analysis}
\begin{figure}[!ht]
	\centering
    \includegraphics[width=.95\columnwidth]{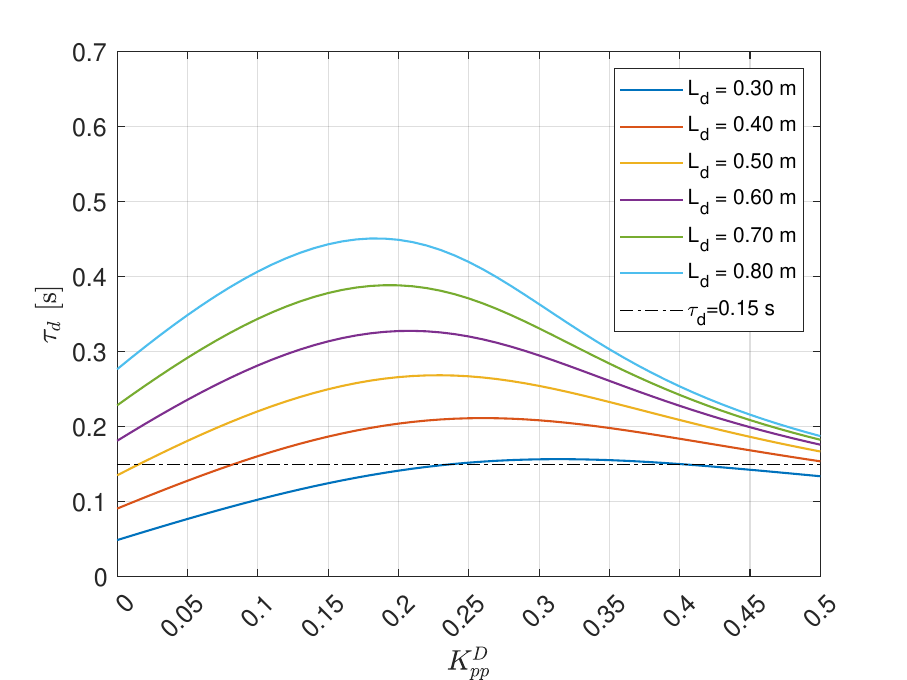}
	\caption{Critical $ \steerdelay $ as a function of the derivative gain in the straight scenario with $\wheelbase=0.26\,\si{\meter}$, $\vel=1\,\si{\meter\per\second}$ and varying lookahead distance. For intervals of $\ppgainder$ corresponding to which the critical delay is bigger than the actual static delay $\steerdelay=0.15\,\si{\second}$, the closed loop system is stable.}
	\label{fig:wm-analysis-pd-straight}
\end{figure}

\begin{figure}[!ht]
	\centering
    \includegraphics[width=.95\columnwidth]{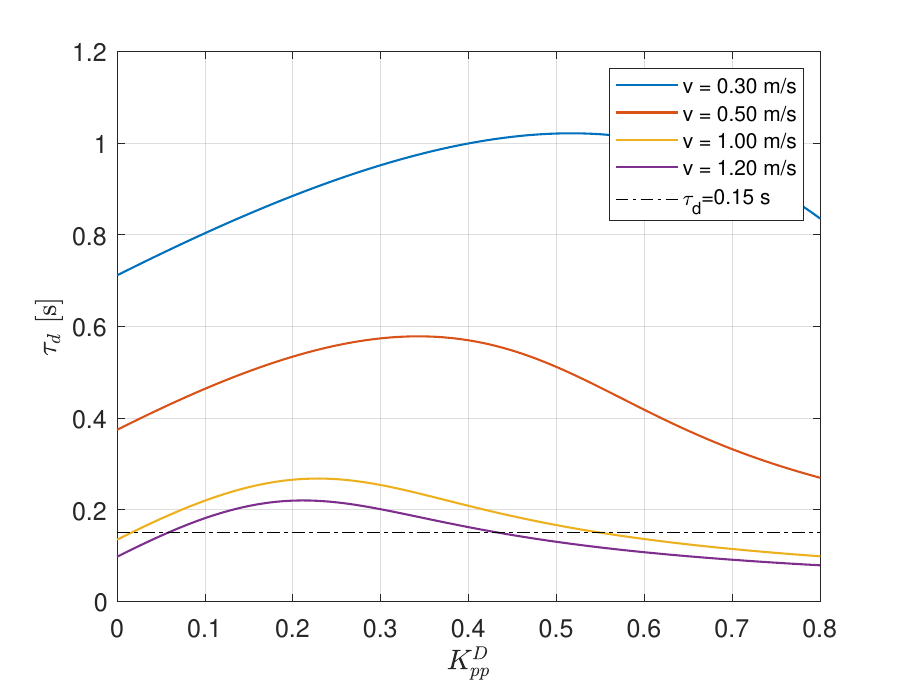}
    	\caption{Critical $ \steerdelay $ as a function of the derivative gain in the straight scenario with $\wheelbase=0.26\,\si{\meter}$, $\lookahead=0.5\,\si{\meter}$ and varying velocity. For intervals of $\ppgainder$ corresponding to which the critical delay is bigger than the actual static delay $\steerdelay=0.15\,\si{\second}$, the closed loop system is stable.}
	\label{fig:wm-analysis-pd-straight-vel}
\end{figure}

In this section we describe the tuning procedure adopted for the PP-based controller.
Regarding the PP-VR generator, practical considerations on the specific scenario suggest the values $\velmax=1\,\si{\meter\per\second}$ and $\lataccmax=0.4\,\si{\meter\per\second\squared}$ as  meaningful parameters in highlighting the the effectiveness of the proposed strategy.
Recalling the WM direct method described in Sec.\ref{sec:delay-analysis}, we refer to \eqref{eq:WM-poly-straight} and \eqref{eq:WM-poly-straight-der} to determine the stability of the delayed closed loop system. The outcome of this procedure, namely the critical delay, is the static delay component up to which the closed loop system is stable for fixed control parameters. In Fig.\ref{fig:wm-analysis-pd-straight} the plot of the critical delay as a function of $ \ppgainder $ for varying lookahead distance and fixed velocity of $1\,\si{\meter\per\second}$ is shown, while in Fig.\ref{fig:wm-analysis-pd-straight-vel} the critical delay for a fixed lookahead distance of $0.5\si{\meter}$ and varying velocity is shown. If the critical delay is above the actual static delay of the system, i.e. $\steerdelay=0.15\,\si{\second}$, the stability of the linearized closed loop system is ensured. \\
It is interesting to notice how too much derivative action, as well as no derivative derivative action, may lead the system to instability. High velocities and small $\lookahead$ lead to unstable systems.

A reasonable choice for $\lookahead$ is imposing it to be smaller than the smallest radius of the reference path, namely $\lookahead < \min_s{\pathradius(s)} = 0.65\,\si{\meter}$, in order to avoid the \emph{``cutting-corners"} effect. According to the just mentioned discussion on Fig. \ref{fig:wm-analysis-pd-straight}, for fixed $\lookahead$, there exists a $\ppgainder$ value for which the critical delay is maximized, suggesting a natural tuning procedure for $\ppgainder$.
The final calibration of the PP requires balancing the control performance and the accuracy of the CNN estimate: for this reason, the adopted value of $\ppgainder$ will be discussed at the end of the following subsection, after having outlined the CNN hyperparameters tuning procedure.




\subsection{CNN Hyperparameters Tuning and Performance Evaluation}

In this section, we address the strategy employed for tuning the hyperparameters of the CNN and validating its performance. The hyperparameter tuning process involves the standard approach of assessing various CNN hyperparameter combinations on a simulated \emph{test dataset} previously unseen. We propose a validate of the tuned network on both a simulated test dataset and a real experimental test dataset: this dual evaluation aims to gauge the general accuracy of the CNN in real-world scenarios and quantify the Sim2Real gap when transitioning from a simulated to a real environment. Indeed, two perfectly aligned datasets, in terms of the position and orientation of the vehicle, are constructed. One dataset is collected in the laboratory, and the other is simulated. Practically, the datasets are generated by properly navigating the car within the laboratory, utilizing the logged pose of the vehicle to create an equivalent dataset in simulation.
The subsequent sections provide a more comprehensive description of the processes involved in dataset generation and hyperparameter tuning.

\subsubsection*{Experimental Test Dataset Generation}
\begin{figure}
    \centering
    \makebox[\linewidth][c]{\includegraphics[width=0.996\linewidth]{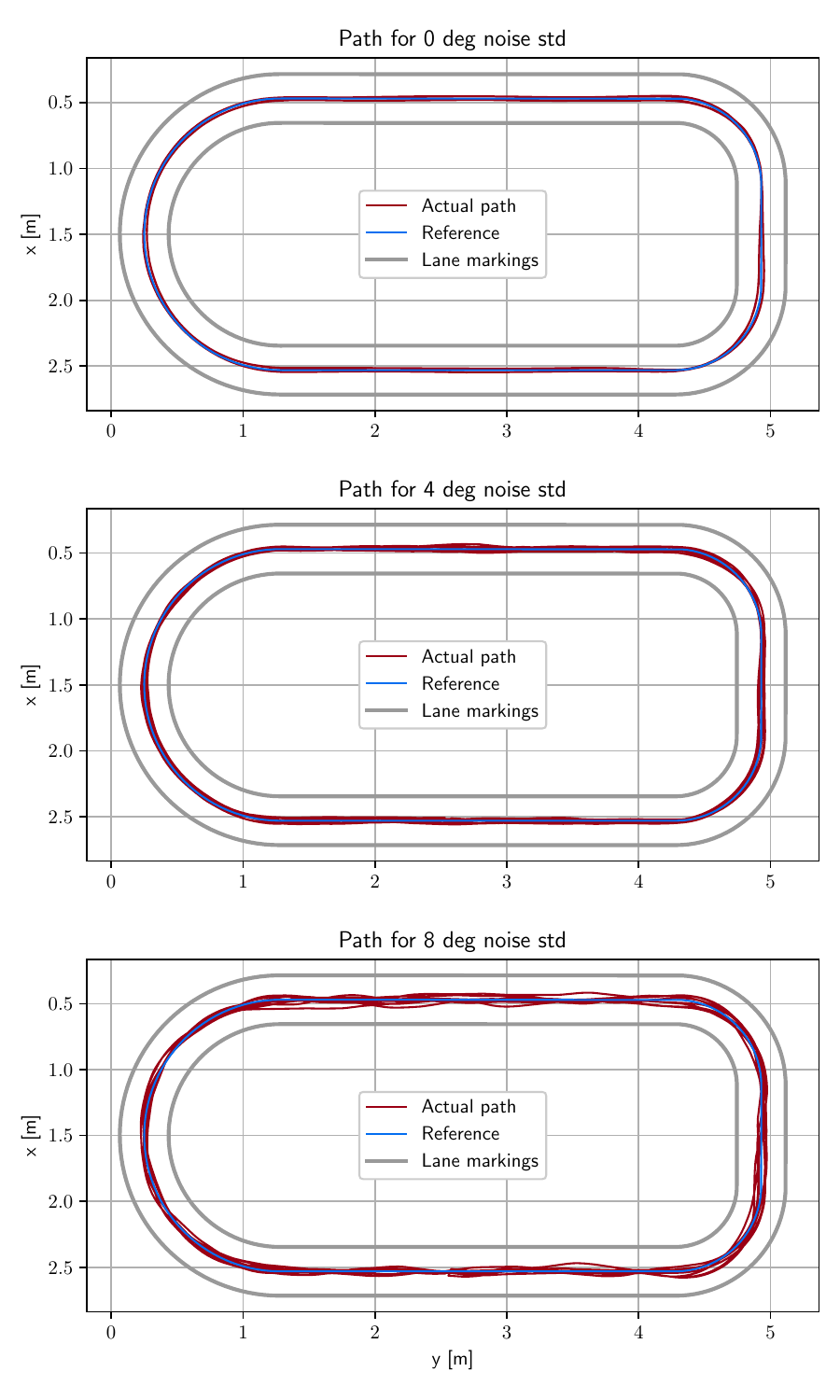}}
    \caption{Selection of experimental test datasets paths.}
    \label{fig:dataset_analysis1}
\end{figure}

To collect real-world data for evaluating the performance of the estimation system, we used a similar method as the one used to generate data for the training datasets (\ref{sec:data_collection}). The true position and orientation of the vehicle were generated using the Vicon system. Since manually moving the car for each sample would be infeasible due to time constraints, we generated image-position pairs by driving the car around the track using a basic PP controller with feedback from the Vicon system. The steering input was periodically perturbed with noise sampled from a Gaussian distribution, with eight different levels of noise applied: $0, 2, 4, 6, 8, 10, 12, 14$ degrees of standard deviation. This allowed us to evaluate the car's performance not only in perfect center-line tracking conditions but also in sub-optimal testing scenarios. For each noise level, the car performed 4 laps around the track clockwise and 4 laps anticlockwise, with a constant speed of $0.3\,\si{\meter\per\second}$. In this study, the LHEs were calculated offline using the method outlined in Section \ref{subsec:geomform}. A range of 8 values for the lookahead distance were tested: $0.2, 0.3, 0.4, 0.5, 0.6, 0.7, 0.8, 0.9$ meters. Figure \ref{fig:dataset_analysis1} depict the actual path taken by the vehicle compared to the center-line reference path. As previously anticipated, to evaluate the generalizability of the proposed strategy in real-world scenarios, the vehicle's positions and orientations were logged during each real-world run and used to generate corresponding evaluation datasets in a simulated environment. 

\subsubsection*{CNN hyperparameters tuning}
The training phase outlined in Section \ref{sec:nn_training}, along with the procedures for generating training datasets detailed in Sections \ref{sec:data_collection} and \ref{sec:dataset_gene}, involves numerous parameters that impact the CNN estimation performance. To enhance the effectiveness of hyperparameter selection beyond initial trial-and-error attempts, a systematic grid exploration of the hyperparameter space was conducted.
This exploration involved repeating the training process outlined in Section \ref{sec:nn_training} for each parameter combination. Evaluation of estimation performance was then performed on the simulated dataset generated using the procedure described in the previous section. The compact architecture of the CNN facilitated the training of a large number of models, a task typically deemed impractical. Table \ref{tab:hypertuning} provides an overview of the considered parameters, their respective value ranges, and the optimal values determined through the grid search. Optimal configurations were obtained by assessing the standard deviation of the CNN estimates against the ground truth of LHE across all samples in the simulated test datasets.

\begin{figure}
    \centering
    \subfloat{\makebox[\linewidth][c]{\includegraphics[width=0.996\linewidth]{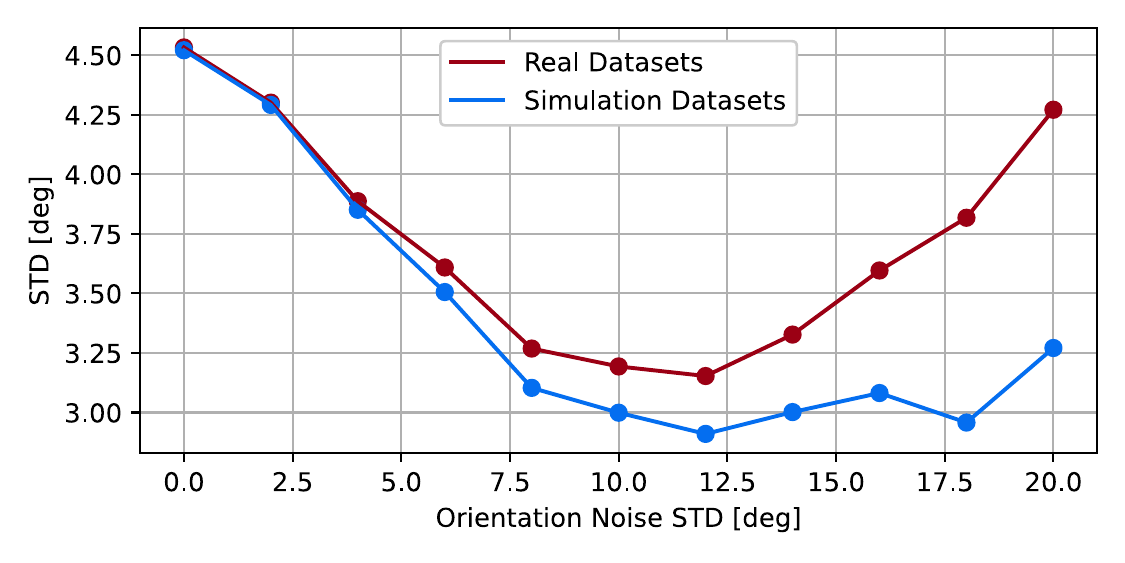}}}\hfill
    \subfloat{\makebox[\linewidth][c]{\includegraphics[width=0.996\linewidth]{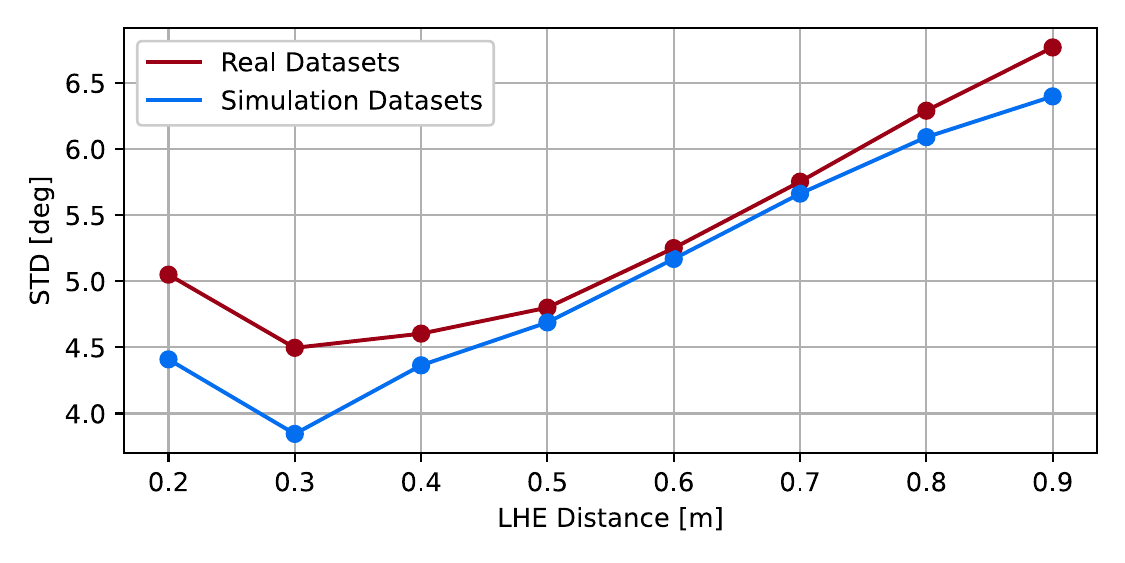}}}
    \caption{\emph{CNN hyperparameters tuning}: STD for orientation noise injected into the training dataset ($\sigma_\head$) and for LHE distances.}
    \label{fig:ds_parameters}
\end{figure}

Supporting the Sim2Real approach, it is noteworthy that the optimization results obtained on real datasets and their simulated counterparts exhibit closely matched values for the majority of parameters. Figure \ref{fig:ds_parameters} illustrates this alignment by presenting the standard deviations of the LHE as a function of the orientation noise injected into the training dataset (referring to $\sigma_\head$ of Section \ref{sec:data_collection}). Additionally, the figure displays LHE standard deviations as a function of LHE distance. In both cases, the optimal values identified in the simulation perfectly correspond to those observed in the real datasets.

\begin{table}
    \centering
    \begin{tabular}{lccc}
        \toprule
        \textbf{Description} & \textbf{Range} & \textbf{Opt. Sim} & \textbf{Opt. Real}\\
        \midrule
        LHE distance [m] &              [0.2, 0.9]      & 0.3   &  0.3  \\
        Dropout probability &           [0, 0.8]        & 0.2   &  0.3  \\
        Orientation noise $\sigma_\head$ [deg]&   [0, 20]         & 12    &  12   \\
        Lateral noise $\sigma_L$ [m]&   [0, 0.1]        & 0.06  &  0.06 \\
        L2 regularization &             [1e-1, 1e-4]    & 1e-3  &  1e-3 \\
        Learning rate &                 [1e-1, 1e-5]    & 3e-3  &  3e-3 \\
        Epochs &                        [50, 500]       & 500   &  300  \\
        Bottom crop percentage &        [0.4, 0.9]      & 0.8   &  0.8  \\
        Blur kernel size &              [0, 7]          & 3     &  3    \\
        Image Gaussian noise std &      [0, 160]        & 80    &  80   \\
        \bottomrule
    \end{tabular}
    \caption{Hyperparameters optimal tuning} \label{tab:hypertuning}
\end{table}

\begin{figure}
    \centering
    \makebox[\linewidth][c]{\includegraphics[width=0.996\linewidth]{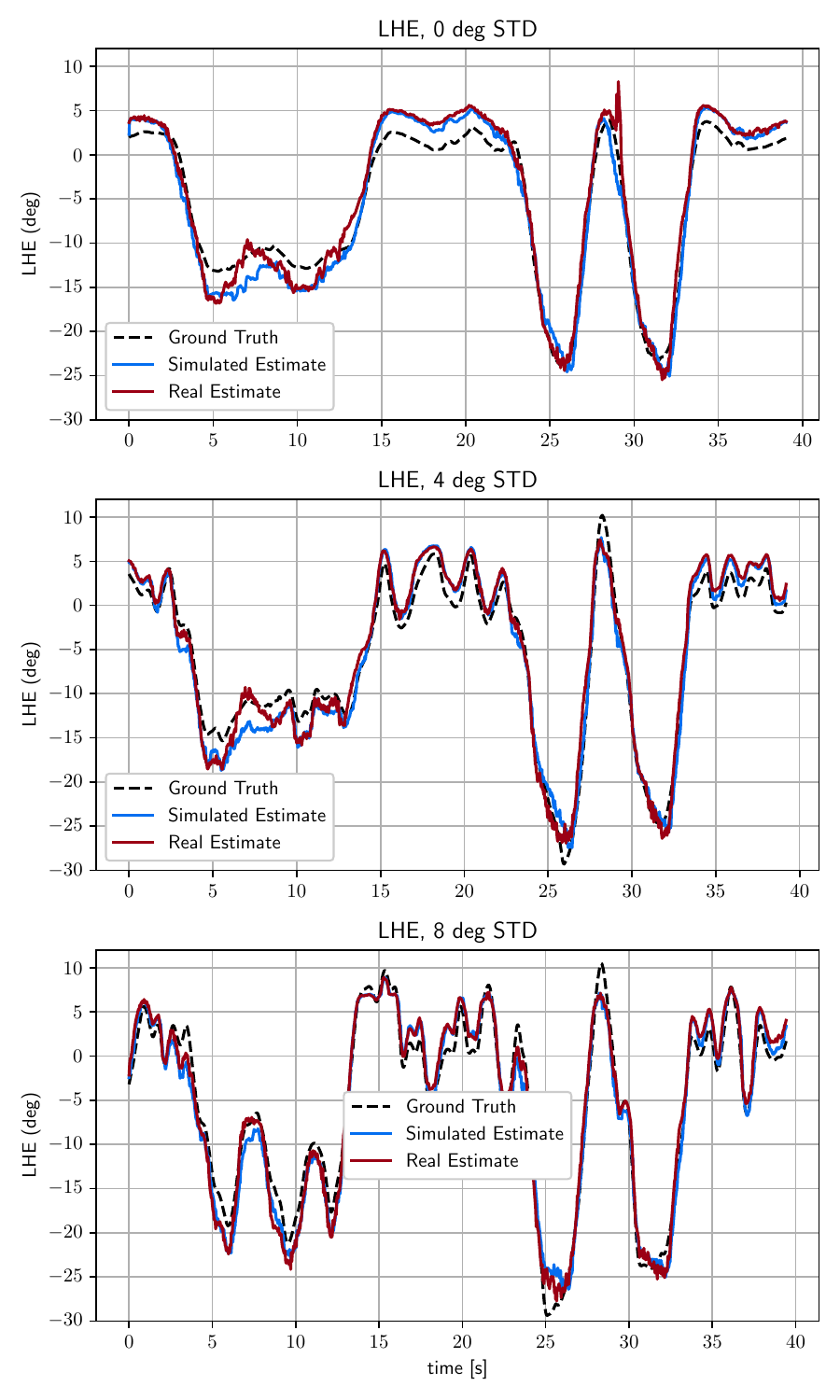}}
    \caption{LHE comparison on selection of test datasets}
    \label{fig:lookahead_comp}
\end{figure}

From a broader perspective of general performance evaluation, Figure \ref{fig:lookahead_comp} provides a comparison of the CNN estimate on three laps of the real test datasets and their simulated counterparts. The figure clearly demonstrates the CNN's high accuracy even in noisy scenarios. Notably, the Sim2Real gap is minimal, as evidenced by the nearly perfect overlap between the real and simulated traces. 
Therefore, the final tuning was selected by balancing the two conflicting needs of the control and the CNN estimation strategy: on the one hand, the first exhibits better stability performance with high values for $\lookahead$, while, on the other hand, the CNN performs better with small $\lookahead$. $\lookahead =0.5$ has been thus adopted, and it is worth noting that the estimation performance of CNN (Fig. \ref{fig:ds_parameters}) does not deteriorate significantly.
According to the rationale proposed in the previous subsection $\ppgainder=0.2$ has been adopted. The resulting best hyperparameters corresponding to this new lookahead distance value are reported in Table \ref{tab:hypertuning5}. 

\begin{table}
    \centering
    \begin{tabular}{lccc}
        \toprule
        \textbf{Description} & \textbf{Range} & \textbf{Opt. Sim} & \textbf{Opt. Real}\\
        \midrule
        Dropout probability &           [0, 0.8]        & 0.3   &  0.2  \\
        Orientation noise $\sigma_\head$ [deg]&   [0, 20]         & 12    &  12   \\
        Lateral noise $\sigma_L$ [m]&   [0, 0.1]        & 0.06  &  0.06 \\
        L2 regularization &             [1e-1, 1e-4]    & 1e-2  &  1e-2 \\
        Learning rate &                 [1e-1, 1e-5]    & 3e-3  &  3e-3 \\
        Epochs &                        [50, 500]       & 300   &  300  \\
        Bottom crop percentage &        [0.4, 0.9]      & 0.8   &  0.8  \\
        Blur kernel size &              [0, 7]          & 3     &  3    \\
        Image Gaussian noise std &      [0, 160]        & 80    &  80   \\
        \bottomrule
    \end{tabular}
    \caption{Best hyperparameters tuning for fixed lookahead distance equal to $0.5\,\si{\meter}$} \label{tab:hypertuning5}
\end{table}

\subsection{Performance on test track}
This section delves into the results of the complete VbLKS proposed in this article.

\begin{figure}[!ht]
    \centering
    \includegraphics[width=\columnwidth]{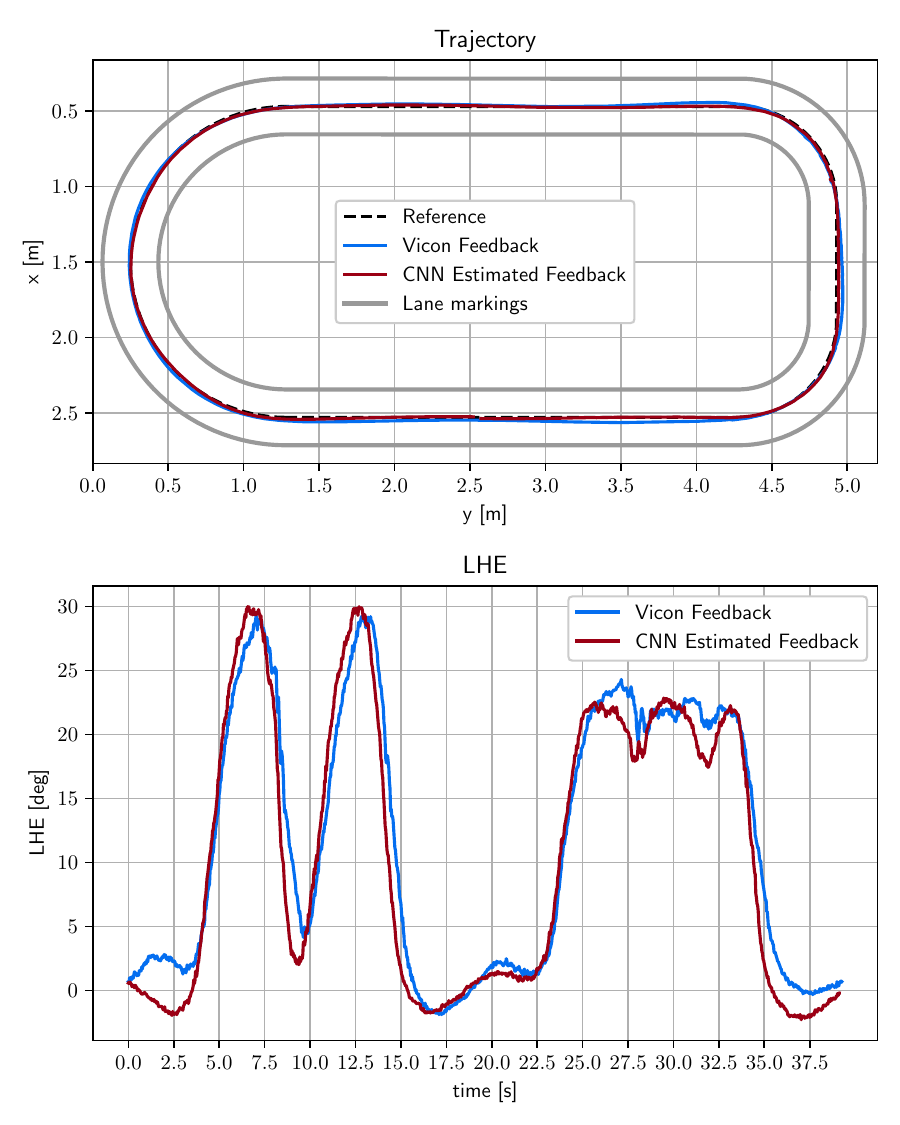}
    \caption{Path along a full lap in counterclockwise direction. The figure compares a solution with feedback from the proposed CNN and another with feedback from the Vicon localization system. Both paths refer to a standard PP controller at fixed velocity of $0.3\,\si{\meter\per\second}$ and lookahead distance of $0.5\,\si{\meter}$.}
    \label{fig:sparcs-comparison}
\end{figure}

Before discussing the results related to the full VbLKS, it is meaningful to test the CNN in a simplified but closed-loop configuration, obtained under the following conditions: given the fixed lookahead distance $\lookahead$ equal to $0.5\,\si{\meter}$, the LHE is estimated by means of the proposed CNN trained in the proposed Sim2Real framework using the optimal parameters, while the PP-D controller is adopted with $\ppgainder=0$ and the velocity is kept constant at $v=0.3\,\si{\meter\per\second}$. 
As a benchmark for comparison, this configuration is compared to another one in which the LHE, fed to the same PP controller, is computed by means of the Vicon localization system. The path and the LHE estimate, depicted in Fig.\ref{fig:sparcs-comparison}, show an almost identical tracking layout, thus providing evidence to the proposed GPS-denied approach.

We now present results referring to the proposed full VbLKS. In this strategy, for fixed lookahead distance $\lookahead$, the LHE is estimated by means of the proposed CNN which is trained using the optimal parameters in the proposed Sim2Real framework. Based on this, the PP-D controller and PP-VR generator are adopted and tuned according to the procedure outlined in Sec.\ref{subsec:tuning-analysis}.

\begin{figure}[!ht]
    \centering
    \includegraphics[width=\columnwidth,trim={1cm 1cm 1cm 1cm},clip]{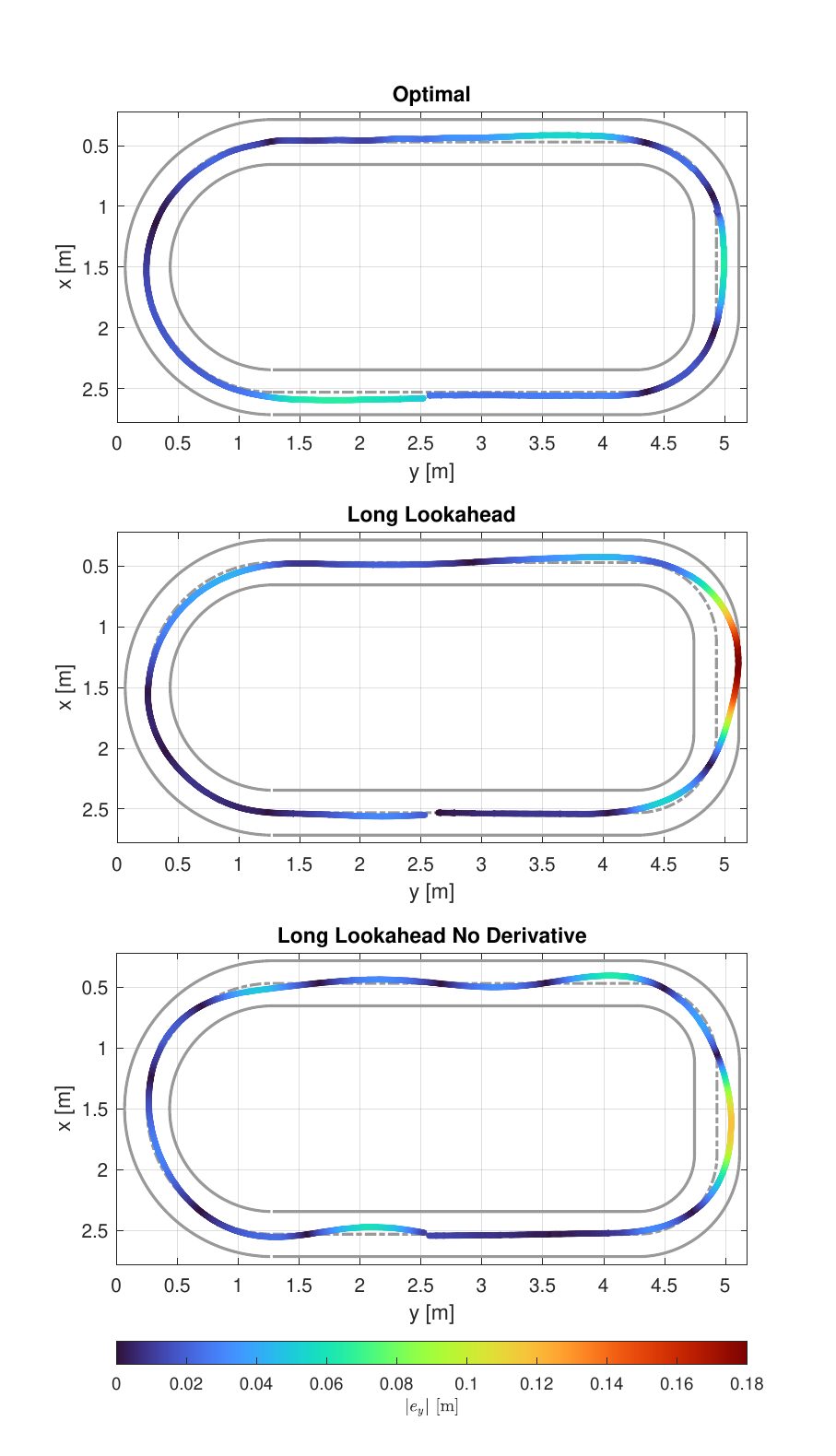}
    \caption{Performance of the proposed VbLKS on the test track}
    \label{fig:sparcs-performance-path}
\end{figure}

\begin{figure}[!ht]
    \centering
    \includegraphics[width=\columnwidth,trim={0cm 0.5cm 0.5cm 0cm},clip]{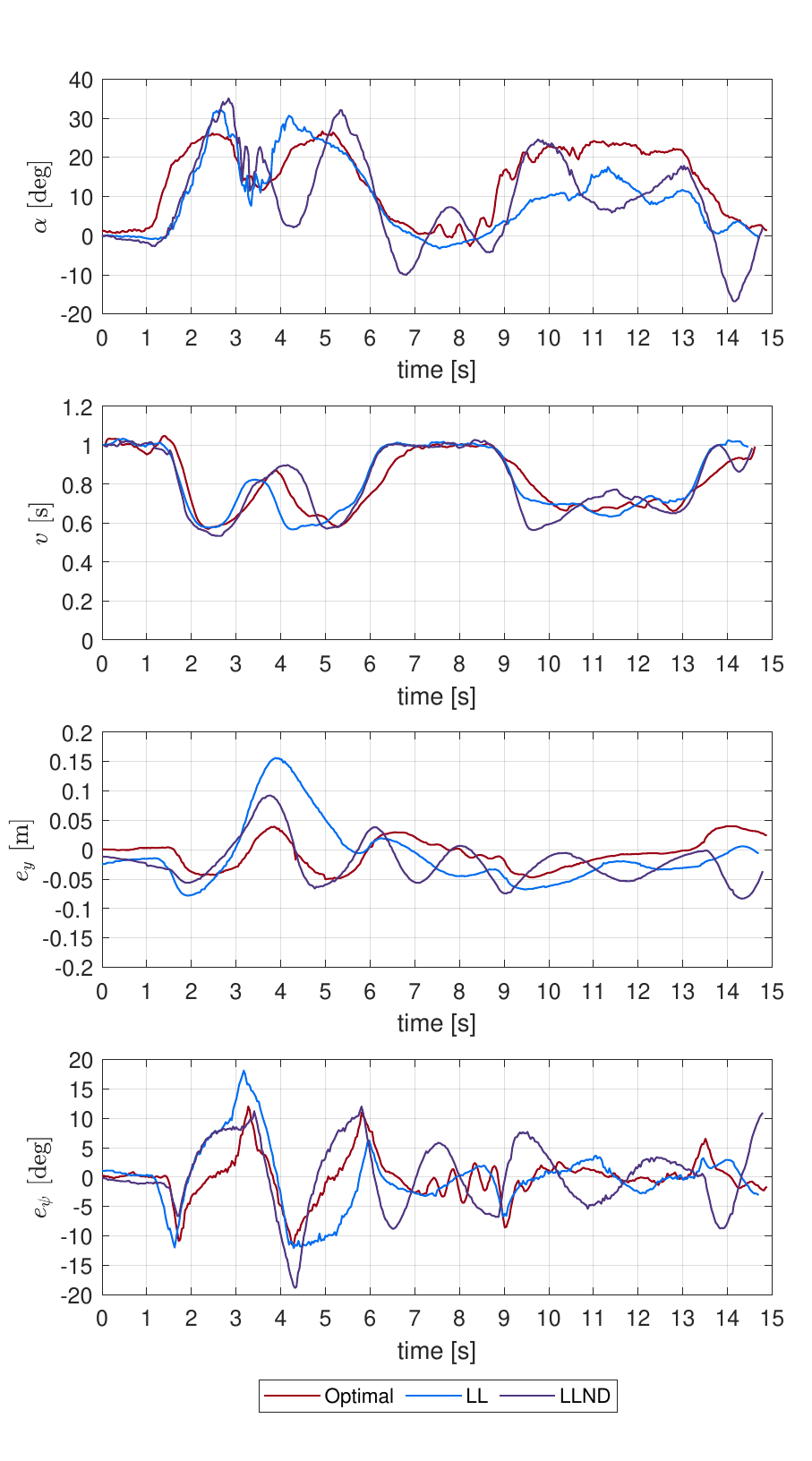}
    \caption{Performance of the proposed VbLKS on the laboratory test track. The first tuning outperforms the others by means of both $\laterrmax$ and $\headerrmax$}
    \label{fig:sparcs-performance-full}
\end{figure}

The efficacy of the proposed strategy is highlighted using three PP-D controller configurations. Besides the \textit{Optimal Configuration} (OC), we consider two other configurations related to a longer lookahead distance in order to introduce effective comparison between the presence and absence of the derivative action. Such a comparison would not be possible in the OC since the absence of the derivative term would make the control system unstable.\\
The following configurations are then considered:
\begin{itemize}
    \item \textit{Optimal Configuration} (OC): it is introduced in Sec.\ref{subsec:tuning-analysis} and it is characterized by $\lookahead=0.5\,\si{\meter}$ and $\ppgainder=0.2$
    \item \textit{Long Lookahead Configuration} (LL-C): a longer $\lookahead$, i.e. $\lookahead=0.8\,\si{\meter}$, is adopted and the corresponding optimal $\ppgainder$ is chosen equal to $0.18$ according to the previous stability analysis.
    \item \textit{Long Lookahead No Derivative Configuration} (LLND-C): this configuration differs from LL-C only for the value of  $\ppgainder$, which is set to $0$.
\end{itemize}
Fig.\ref{fig:sparcs-performance-path} depicts the performance of each configuration respectively. The following indexes on the lateral and heading errors are considered to evaluate performance \cite{Rokonuzzaman2021}:
\begin{align}\label{eq:performance-metrics}
	\laterrmax &= \max_{t\in[0,T]}{\left|\laterr(t)\right|} & \text{\scshape maximum lateral error}\nonumber\\
	\headerrmax &= \max_{t\in[0,T]}{\left|\headerr(t)\right|} & \text{\scshape maximum heading error}\nonumber		
\end{align}
These indices can be extracted from Fig.\ref{fig:sparcs-performance-full}.
The first configuration outperforms the others in terms of both maximum lateral and heading error. Indeed, thanks to the damping effect of the added derivative action, the ``cutting-corner'' phenomenon can be mitigated by decreasing the lookahead distance while retaining tracking performance along straight sections.
The effect of the derivative action can be appreciated in the following two configurations. In the LLND-C it can be noticed how just increasing the lookahead distance is not sufficient in damping oscillations along straight sections due to the presence of delay in the actuation mechanism. Beyond the fact of giving rise to the ``cutting-corner'' phenomenon along tight curves. Introducing the derivative action (LL-C) allows to damp oscillations but is not effective in mitigating the ``cutting-corner'' phenomenon.\\
Eventually, variations to the proposed calibrations were evaluated in the experimental phase without significant advantages.

\section{Conclusions}
In this article, the development of a Sim2Real GPS-denied VbLKS for a 1:10-scale AV has been presented. In this system, the training of a CNN has been carried out within a Sim2Real framework and a tailored PP-based control strategy has been employed to mitigate delays in the steering actuation mechanism of the vehicle. Notably, the VbLKS has been proven to effectively address the reality gap in a GPS-denied reference scenario, while demonstrating real-time functionality on low-level embedded hardware.
The effectiveness of the proposed VbLKS architecture has been demonstrated in a laboratory setting and excelled in the BFMC 2022 structured reference scenario.\vfill

\bibliographystyle{IEEEtranN}
\bibliography{./bibliography}

\begin{thebibliography}{47}
\providecommand{\natexlab}[1]{#1}
\providecommand{\url}[1]{#1}
\csname url@samestyle\endcsname
\providecommand{\newblock}{\relax}
\providecommand{\bibinfo}[2]{#2}
\providecommand{\BIBentrySTDinterwordspacing}{\spaceskip=0pt\relax}
\providecommand{\BIBentryALTinterwordstretchfactor}{4}
\providecommand{\BIBentryALTinterwordspacing}{\spaceskip=\fontdimen2\font plus
\BIBentryALTinterwordstretchfactor\fontdimen3\font minus \fontdimen4\font\relax}
\providecommand{\BIBforeignlanguage}[2]{{%
\expandafter\ifx\csname l@#1\endcsname\relax
\typeout{** WARNING: IEEEtranN.bst: No hyphenation pattern has been}%
\typeout{** loaded for the language `#1'. Using the pattern for}%
\typeout{** the default language instead.}%
\else
\language=\csname l@#1\endcsname
\fi
#2}}
\providecommand{\BIBdecl}{\relax}
\BIBdecl

\bibitem[Chen et~al.(2023)Chen, Li, Huang, Li, Xing, Tian, Li, Hu, Na, Li, Teng, Lv, Wang, Cao, Zheng, and Wang]{Chen2022}
L.~Chen, Y.~Li, C.~Huang, B.~Li, Y.~Xing, D.~Tian, L.~Li, Z.~Hu, X.~Na, Z.~Li, S.~Teng, C.~Lv, J.~Wang, D.~Cao, N.~Zheng, and F.-Y. Wang, ``Milestones in autonomous driving and intelligent vehicles: Survey of surveys,'' \emph{IEEE Transactions on Intelligent Vehicles}, vol.~8, no.~2, pp. 1046--1056, 2023.

\bibitem[of~Automotive~Engineers(2021)]{SAEJ3016}
\BIBentryALTinterwordspacing
S.~of~Automotive~Engineers, \emph{Taxonomy and Definitions for Terms Related to Driving Automation Systems for On-Road Motor Vehicles}.\hskip 1em plus 0.5em minus 0.4em\relax SAE International, 2021. [Online]. Available: \url{https://www.sae.org/standards/content/j3016\_202104/}
\BIBentrySTDinterwordspacing

\bibitem[Buehler et~al.(2009)Buehler, Iagnemma, and Singh]{DARPA2009}
M.~Buehler, K.~Iagnemma, and S.~Singh, \emph{The DARPA Urban Challenge: Autonomous Vehicles in City Traffic}, 1st~ed.\hskip 1em plus 0.5em minus 0.4em\relax Springer Publishing Company, Incorporated, 2009.

\bibitem[Campbell(2007)]{Campbell2007}
S.~F. Campbell, ``Steering control of an autonomous ground vehicle with application to the darpa urban challenge,'' 2007.

\bibitem[Kalapos et~al.(2020)Kalapos, Gór, Moni, and Harmati]{Kalapos2020}
A.~Kalapos, C.~Gór, R.~Moni, and I.~Harmati, ``Sim-to-real reinforcement learning applied to end-to-end vehicle control,'' in \emph{2020 23rd International Symposium on Measurement and Control in Robotics (ISMCR)}, 2020, pp. 1--6.

\bibitem[Almási et~al.(2020)Almási, Moni, and Gyires-Tóth]{Almasi2020}
P.~Almási, R.~Moni, and B.~Gyires-Tóth, ``Robust reinforcement learning-based autonomous driving agent for simulation and real world,'' in \emph{2020 International Joint Conference on Neural Networks (IJCNN)}, 2020, pp. 1--8.

\bibitem[Karouach and Ivanov(2016)]{Cars1}
I.~Karouach and S.~Ivanov, ``Lane detection and following approach in self-driving miniature vehicles,'' 2016.

\bibitem[Sun et~al.(2019)Sun, Zheng, Qiao, Liu, Lin, and Bräunl]{Cars2}
\BIBentryALTinterwordspacing
S.~Sun, J.~Zheng, Z.~Qiao, S.~Liu, Z.~Lin, and T.~Bräunl, ``The architecture of a driverless robot car based on eyebot system,'' \emph{Journal of Physics: Conference Series}, vol. 1267, no.~1, p. 012099, jul 2019. [Online]. Available: \url{https://dx.doi.org/10.1088/1742-6596/1267/1/012099}
\BIBentrySTDinterwordspacing

\bibitem[Palma et~al.(2020)Palma, Bonilla, and Grande]{Cars3}
J.~A.~B. Palma, M.~N.~I. Bonilla, and R.~E. Grande, ``Lane line detection computer vision system applied to a scale autonomos car: Automodelcar,'' in \emph{2020 17th International Conference on Electrical Engineering, Computing Science and Automatic Control (CCE)}, 2020, pp. 1--6.

\bibitem[Hu et~al.(2023)Hu, Li, Huang, Tang, Huai, and Chen]{Hu2023}
X.~Hu, S.~Li, T.~Huang, B.~Tang, R.~Huai, and L.~Chen, ``How simulation helps autonomous driving:a survey of sim2real, digital twins, and parallel intelligence,'' 2023.

\bibitem[Kilyen et~al.(2021)Kilyen, Lemnariu, Mois, Chen, Morris, and Muntean]{BFMC-ITSS}
N.~A. Kilyen, R.~F. Lemnariu, G.~D. Mois, Y.~Chen, B.~T. Morris, and I.~Muntean, ``The ieee itss and bosch future mobility challenge: A hands-on start to autonomous driving [technical activities],'' \emph{IEEE Intelligent Transportation Systems Magazine}, vol.~13, no.~3, pp. 276--282, 2021.

\bibitem[Paden et~al.(2016)Paden, Cap, Yong, Yershov, and Frazzoli]{Paden2016}
\BIBentryALTinterwordspacing
B.~Paden, M.~Cap, S.~Z. Yong, D.~Yershov, and E.~Frazzoli, ``A survey of motion planning and control techniques for self-driving urban vehicles,'' 2016. [Online]. Available: \url{https://arxiv.org/abs/1604.07446}
\BIBentrySTDinterwordspacing

\bibitem[Rokonuzzaman et~al.(2021)Rokonuzzaman, Mohajer, Nahavandi, and Mohamed]{Rokonuzzaman2021}
\BIBentryALTinterwordspacing
M.~Rokonuzzaman, N.~Mohajer, S.~Nahavandi, and S.~Mohamed, ``Review and performance evaluation of path tracking controllers of autonomous vehicles,'' \emph{IET Intelligent Transport Systems}, vol.~15, no.~5, pp. 646--670, 2021. [Online]. Available: \url{https://ietresearch.onlinelibrary.wiley.com/doi/abs/10.1049/itr2.12051}
\BIBentrySTDinterwordspacing

\bibitem[Kumar et~al.(2022)Kumar, Saini, Pandey, Agarwal, Agrawal, and Agarwal]{Kumar2022}
\BIBentryALTinterwordspacing
A.~Kumar, T.~Saini, P.~B. Pandey, A.~Agarwal, A.~Agrawal, and B.~Agarwal, ``Vision-based outdoor navigation of self-driving car using lane detection,'' \emph{International Journal of Information Technology}, vol.~14, no.~1, pp. 215--227, Feb 2022. [Online]. Available: \url{https://doi.org/10.1007/s41870-021-00747-2}
\BIBentrySTDinterwordspacing

\bibitem[Diab et~al.(2020)Diab, Abbas, Ammar, and Shalaby]{Diab2020a}
M.~K. Diab, A.~N. Abbas, H.~H. Ammar, and R.~Shalaby, ``Experimental lane keeping assist for an autonomous vehicle based on optimal pid controller,'' in \emph{2020 2nd Novel Intelligent and Leading Emerging Sciences Conference (NILES)}, 2020, pp. 486--491.

\bibitem[Cantas and Guvenc(2018)]{Cantas2018}
\BIBentryALTinterwordspacing
M.~R. Cantas and L.~Guvenc, ``Camera based automated lane keeping application complemented by gps localization based path following,'' in \emph{WCX World Congress Experience}.\hskip 1em plus 0.5em minus 0.4em\relax SAE International, apr 2018. [Online]. Available: \url{https://doi.org/10.4271/2018-01-0608}
\BIBentrySTDinterwordspacing

\bibitem[Kuo et~al.(2019)Kuo, Lu, and Yang]{Kuo2019}
\BIBentryALTinterwordspacing
C.~Kuo, Y.~Lu, and S.~Yang, ``On the image sensor processing for lane detection and control in vehicle lane keeping systems,'' \emph{Sensors}, vol.~19, no.~7, 2019. [Online]. Available: \url{https://www.mdpi.com/1424-8220/19/7/1665}
\BIBentrySTDinterwordspacing

\bibitem[Ko~Htet et~al.(2015)Ko~Htet, Kiong, and Xinxin]{Htet2015}
K.~K. Ko~Htet, T.~K. Kiong, and D.~Xinxin, ``Comprehensive lane keeping system with mono camera,'' in \emph{2015 10th Asian Control Conference (ASCC)}, 2015, pp. 1--6.

\bibitem[Satria et~al.(2022)Satria, Indriawati, Widjiantoro, Hija, and Nurhadi]{Satria2022}
M.~A. Satria, K.~Indriawati, B.~L. Widjiantoro, A.~I. Hija, and H.~Nurhadi, ``Lane keeping control using nonlinear model predictive control on constant speed autonomous car,'' in \emph{2022 IEEE International Conference on Cybernetics and Computational Intelligence (CyberneticsCom)}, 2022, pp. 12--16.

\bibitem[Schwarting et~al.(2018)Schwarting, Alonso-Mora, and Rus]{Schwarting2018}
\BIBentryALTinterwordspacing
W.~Schwarting, J.~Alonso-Mora, and D.~Rus, ``Planning and decision-making for autonomous vehicles,'' \emph{Annual Review of Control, Robotics, and Autonomous Systems}, vol.~1, no.~1, pp. 187--210, 2018. [Online]. Available: \url{https://doi.org/10.1146/annurev-control-060117-10515}
\BIBentrySTDinterwordspacing

\bibitem[Bojarski et~al.(2017)Bojarski, Yeres, Choromanska, Choromanski, Firner, Jackel, and Muller]{Bojarski2017}
\BIBentryALTinterwordspacing
M.~Bojarski, P.~Yeres, A.~Choromanska, K.~Choromanski, B.~Firner, L.~Jackel, and U.~Muller, ``Explaining how a deep neural network trained with end-to-end learning steers a car,'' 2017. [Online]. Available: \url{https://arxiv.org/abs/1704.07911}
\BIBentrySTDinterwordspacing

\bibitem[Gidado et~al.(2020)Gidado, Chiroma, Aljojo, Abubakar, Popoola, and Al-Garadi]{Gidado2020}
U.~M. Gidado, H.~Chiroma, N.~Aljojo, S.~Abubakar, S.~I. Popoola, and M.~A. Al-Garadi, ``A survey on deep learning for steering angle prediction in autonomous vehicles,'' \emph{IEEE Access}, vol.~8, pp. 163\,797--163\,817, 2020.

\bibitem[Kocić and Jovičić(2021)]{Kocic2021}
J.~Kocić and N.~Jovičić, ``Sim-to-real autonomous vehicle lane keeping using vision,'' in \emph{2021 29th Telecommunications Forum (TELFOR)}, 2021, pp. 1--8.

\bibitem[Yenikaya et~al.(2013)Yenikaya, Yenikaya, and D\"{u}ven]{Yenikaya2013}
\BIBentryALTinterwordspacing
S.~Yenikaya, G.~Yenikaya, and E.~D\"{u}ven, ``Keeping the vehicle on the road: A survey on on-road lane detection systems,'' \emph{ACM Comput. Surv.}, vol.~46, no.~1, jul 2013. [Online]. Available: \url{https://doi.org/10.1145/2522968.2522970}
\BIBentrySTDinterwordspacing

\bibitem[Xing et~al.(2018)Xing, Lv, Chen, Wang, Wang, Cao, Velenis, and Wang]{Xing2018}
Y.~Xing, C.~Lv, L.~Chen, H.~Wang, H.~Wang, D.~Cao, E.~Velenis, and F.-Y. Wang, ``Advances in vision-based lane detection: Algorithms, integration, assessment, and perspectives on acp-based parallel vision,'' \emph{IEEE/CAA Journal of Automatica Sinica}, vol.~5, no.~3, pp. 645--661, 2018.

\bibitem[Huang and Liu(2021)]{Huang2021}
\BIBentryALTinterwordspacing
Q.~Huang and J.~Liu, ``Practical limitations of lane detection algorithm based on hough transform in challenging scenarios,'' \emph{International Journal of Advanced Robotic Systems}, vol.~18, no.~2, p. 17298814211008752, 2021. [Online]. Available: \url{https://doi.org/10.1177/17298814211008752}
\BIBentrySTDinterwordspacing

\bibitem[Zakaria et~al.(2023)Zakaria, Shapiai, Ghani, Yassin, Ibrahim, and Wahid]{Zakaria2022}
N.~J. Zakaria, M.~I. Shapiai, R.~A. Ghani, M.~N.~M. Yassin, M.~Z. Ibrahim, and N.~Wahid, ``Lane detection in autonomous vehicles: A systematic review,'' \emph{IEEE Access}, vol.~11, pp. 3729--3765, 2023.

\bibitem[Waykole et~al.(2021)Waykole, Shiwakoti, and Stasinopoulos]{Waykole2021}
\BIBentryALTinterwordspacing
S.~Waykole, N.~Shiwakoti, and P.~Stasinopoulos, ``Review on lane detection and tracking algorithms of advanced driver assistance system,'' \emph{Sustainability}, vol.~13, no.~20, 2021. [Online]. Available: \url{https://www.mdpi.com/2071-1050/13/20/11417}
\BIBentrySTDinterwordspacing

\bibitem[Ortegon-Sarmiento et~al.(2022)Ortegon-Sarmiento, Kelouwani, Alam, Uribe-Quevedo, Amamou, Paderewski-Rodriguez, and Gutierrez-Vela]{Sarmiento2022}
\BIBentryALTinterwordspacing
T.~Ortegon-Sarmiento, S.~Kelouwani, M.~Z. Alam, A.~Uribe-Quevedo, A.~Amamou, P.~Paderewski-Rodriguez, and F.~Gutierrez-Vela, ``Analyzing performance effects of neural networks applied to lane recognition under various environmental driving conditions,'' \emph{World Electric Vehicle Journal}, vol.~13, no.~10, 2022. [Online]. Available: \url{https://www.mdpi.com/2032-6653/13/10/191}
\BIBentrySTDinterwordspacing

\bibitem[Coulter(1992)]{Coulter1992}
R.~C. Coulter, ``Implementation of the pure pursuit path tracking algorithm,'' Carnegie Mellon University, Pittsburgh, PA, Tech. Rep. CMU-RI-TR-92-01, January 1992.

\bibitem[Iandola and Keutzer(2017)]{Iandola2017}
\BIBentryALTinterwordspacing
F.~Iandola and K.~Keutzer, ``Small neural nets are beautiful: Enabling embedded systems with small deep-neural-network architectures,'' in \emph{Proceedings of the Twelfth IEEE/ACM/IFIP International Conference on Hardware/Software Codesign and System Synthesis Companion}, ser. CODES '17.\hskip 1em plus 0.5em minus 0.4em\relax New York, NY, USA: Association for Computing Machinery, 2017. [Online]. Available: \url{https://doi.org/10.1145/3125502.3125606}
\BIBentrySTDinterwordspacing

\bibitem[Wang et~al.(2022)Wang, Chen, and Zhu]{Wang2022}
\BIBentryALTinterwordspacing
L.~Wang, Z.~Chen, and W.~Zhu, ``An improved pure pursuit path tracking control method based on heading error rate,'' \emph{Industrial Robot: the international journal of robotics research and application}, vol.~49, no.~5, pp. 973--980, Jan 2022. [Online]. Available: \url{https://doi.org/10.1108/IR-11-2021-0257}
\BIBentrySTDinterwordspacing

\bibitem[Canny(1986)]{canny}
J.~Canny, ``A computational approach to edge detection,'' \emph{Pattern Analysis and Machine Intelligence, IEEE Transactions on}, vol. PAMI-8, pp. 679 -- 698, 12 1986.

\bibitem[Zhong et~al.(2017)Zhong, Zheng, Kang, Li, and Yang]{random_erase}
\BIBentryALTinterwordspacing
Z.~Zhong, L.~Zheng, G.~Kang, S.~Li, and Y.~Yang, ``Random erasing data augmentation,'' 2017. [Online]. Available: \url{https://arxiv.org/abs/1708.04896}
\BIBentrySTDinterwordspacing

\bibitem[Lecun et~al.(2000)Lecun, Haffner, and Bengio]{lecun}
Y.~Lecun, P.~Haffner, and Y.~Bengio, ``Object recognition with gradient-based learning,'' 08 2000.

\bibitem[Ioffe and Szegedy(2015)]{batch_norm}
\BIBentryALTinterwordspacing
S.~Ioffe and C.~Szegedy, ``Batch normalization: Accelerating deep network training by reducing internal covariate shift,'' 2015. [Online]. Available: \url{https://arxiv.org/abs/1502.03167}
\BIBentrySTDinterwordspacing

\bibitem[Srivastava et~al.(2014)Srivastava, Hinton, Krizhevsky, Sutskever, and Salakhutdinov]{dropout}
\BIBentryALTinterwordspacing
N.~Srivastava, G.~Hinton, A.~Krizhevsky, I.~Sutskever, and R.~Salakhutdinov, ``Dropout: A simple way to prevent neural networks from overfitting,'' \emph{Journal of Machine Learning Research}, vol.~15, no.~56, pp. 1929--1958, 2014. [Online]. Available: \url{http://jmlr.org/papers/v15/srivastava14a.html}
\BIBentrySTDinterwordspacing

\bibitem[Snider(2009)]{Snider2009}
J.~M. Snider, ``Automatic steering methods for autonomous automobile path tracking,'' Carnegie Mellon University, Pittsburgh, PA, Tech. Rep. CMU-RI-TR-09-08, February 2009.

\bibitem[Sukhil and Behl(2021)]{Sukhil2021}
\BIBentryALTinterwordspacing
V.~Sukhil and M.~Behl, ``Adaptive lookahead pure-pursuit for autonomous racing,'' \emph{CoRR}, vol. abs/2111.08873, 2021. [Online]. Available: \url{https://arxiv.org/abs/2111.08873}
\BIBentrySTDinterwordspacing

\bibitem[Park et~al.(2004)Park, Deyst, and How]{Park04}
S.~Park, J.~Deyst, and J.~P. How, ``A new nonlinear guidance logic for trajectory tracking,'' in \emph{In Proceedings of the AIAA Guidance, Navigation and Control Conference}, 2004, pp. 2004--4900.

\bibitem[De~Luca et~al.(1998)De~Luca, Oriolo, and Samson]{DeLuca1998}
\BIBentryALTinterwordspacing
A.~De~Luca, G.~Oriolo, and C.~Samson, \emph{Feedback control of a nonholonomic car-like robot}.\hskip 1em plus 0.5em minus 0.4em\relax Berlin, Heidelberg: Springer Berlin Heidelberg, 1998, pp. 171--253. [Online]. Available: \url{https://doi.org/10.1007/BFb0036073}
\BIBentrySTDinterwordspacing

\bibitem[Xu et~al.(2021)Xu, Peng, and Tang]{Xu2021}
S.~Xu, H.~Peng, and Y.~Tang, ``Preview path tracking control with delay compensation for autonomous vehicles,'' \emph{IEEE Transactions on Intelligent Transportation Systems}, vol.~22, no.~5, pp. 2979--2989, 2021.

\bibitem[Walton and Marshall(1987)]{Walton1987}
\BIBentryALTinterwordspacing
K.~Walton and J.~Marshall, ``\BIBforeignlanguage{English}{Direct method for tds stability analysis},'' \emph{\BIBforeignlanguage{English}{IEE Proceedings D (Control Theory and Applications)}}, vol. 134, pp. 101--107(6), March 1987. [Online]. Available: \url{https://digital-library.theiet.org/content/journals/10.1049/ip-d.1987.0018}
\BIBentrySTDinterwordspacing

\bibitem[Sil(2005)]{Silva2005}
\BIBentryALTinterwordspacing
\emph{Preliminary Results for Analyzing Systems with Time Delay}.\hskip 1em plus 0.5em minus 0.4em\relax Boston, MA: Birkh{\"a}user Boston, 2005, pp. 77--107. [Online]. Available: \url{https://doi.org/10.1007/0-8176-4423-75}
\BIBentrySTDinterwordspacing

\bibitem[Vy~Nguyen et~al.(2019)Vy~Nguyen, Bonnet, Boussaada, and Souaiby]{Nguyen2019}
L.~H. Vy~Nguyen, C.~Bonnet, I.~Boussaada, and M.~Souaiby, ``A problematic issue in the walton-marshall method for some neutral delay systems,'' in \emph{2019 IEEE 58th Conference on Decision and Control (CDC)}, 2019, pp. 971--975.

\bibitem[Ollero and Heredia(1995)]{Ollero1995}
A.~Ollero and G.~Heredia, ``Stability analysis of mobile robot path tracking,'' in \emph{Proceedings 1995 IEEE/RSJ International Conference on Intelligent Robots and Systems. Human Robot Interaction and Cooperative Robots}, vol.~3, 1995, pp. 461--466 vol.3.

\bibitem[Filho et~al.(2014)Filho, Wolf, Grassi, and Osório]{Filho2014}
C.~M. Filho, D.~F. Wolf, V.~Grassi, and F.~S. Osório, ``Longitudinal and lateral control for autonomous ground vehicles,'' in \emph{2014 IEEE Intelligent Vehicles Symposium Proceedings}, 2014, pp. 588--593.

\end{thebibliography}

\end{document}